\documentclass[sigplan,screen]{acmart}

\usepackage{subfig}
\usepackage[normalem]{ulem}

\usepackage{xspace}
\newcommand{\projectname}{{\tt SmartMem}\xspace}
\newcommand\nott[1]{\bgroup\let\tt\relax#1\egroup}

\newcommand{\revisioncr}[1]{#1}

\usepackage[]{hyperref}
\usepackage{soul}
\usepackage{ctable}
\usepackage{booktabs}
\usepackage{longtable}
\usepackage{float}
\usepackage{lscape}
\usepackage[T1]{fontenc}
\usepackage{algcompatible}
\usepackage{amsmath} 
\newcommand{\sslash}{\mathbin{/\mkern-6mu/}}
\usepackage{amsfonts}
\usepackage{pifont}
\usepackage{balance}
\usepackage{array}
\newcolumntype{P}[1]{>{\centering\arraybackslash}p{#1}}
\usepackage{listings}
\definecolor{keyword}{HTML}{2F7AFF}
\definecolor{identifier}{HTML}{000000}
\definecolor{commentstyle}{HTML}{5E5E5E}
\lstdefinelanguage{mylanguage}{
  keywords={True, False, return, switch, if, in, while, do, else, case, break, def, for},
  keywordstyle=\color{keyword}\bfseries,
  ndkeywords={class, export, boolean, throw, implements, import, this},
  ndkeywordstyle=\color{darkgray}\bfseries,
  identifierstyle=\color{identifier},
  sensitive=false,
  comment=[l]{\#},
  morecomment=[s]{/*}{*/},
  commentstyle=\color{commentstyle}\ttfamily,
  stringstyle=\color{red}\ttfamily,
  morestring=[b]',
  morestring=[b]"
}
\lstdefinestyle{mystyle}{
    language=mylanguage,
    backgroundcolor=\color{white},
    numberstyle=\tiny\color{black},
    basicstyle=\ttfamily\footnotesize,
    breakatwhitespace=false,
    breaklines=true,
    captionpos=b,
    keepspaces=true,
    numbers=left,
    numbersep=3pt,
    stepnumber=1,
    showspaces=false,
    showstringspaces=false,
    showtabs=false,
    tabsize=2,
    morekeywords={entry},
    frame=tb,
}
\usepackage[ruled,vlined,linesnumbered]{algorithm2e}
\usepackage[normalem]{ulem}
\usepackage{multirow}
\usepackage{multicol}
\usepackage{makecell}
\usepackage{xspace}
\usepackage{enumitem}
\usepackage{lipsum}
\usepackage{ragged2e}
\usepackage{diagbox}
\usepackage{MnSymbol}
\usepackage{mathrsfs}
\definecolor{miao}{RGB}{140,114,143}
\definecolor{cell1}{HTML}{FF5B00}
\definecolor{cell2}{HTML}{ffa200}
\definecolor{cell3}{HTML}{f9da24}
\definecolor{cell4}{HTML}{f9da24}
\definecolor{cell5}{HTML}{cff09e}
\newcommand{\tdim}[1]{#1{\text -}d}
\newcommand{\opname}[1]{\tt #1:}
\newcommand{\optype}[1]{\tt}

\usepackage{colortbl} 

\copyrightyear{2024} 
\acmYear{2024} 
\setcopyright{rightsretained} 
\acmConference[ASPLOS '24]{29th ACM International Conference on
Architectural Support for Programming Languages and Operating Systems,
Volume 3}{April 27-May 1, 2024}{La Jolla, CA, USA}
\acmBooktitle{29th ACM International Conference on Architectural Support for
Programming Languages and Operating Systems, Volume 3 (ASPLOS '24), April
27-May 1, 2024, La Jolla, CA, USA}
\acmDOI{10.1145/3620666.3651384}
\acmISBN{979-8-4007-0386-7/24/04}

\begin{document}

\title{SmartMem: Layout Transformation Elimination and Adaptation for Efficient DNN Execution on Mobile}

\author{Wei Niu}
\email{wniu@uga.edu}
\affiliation{%
  \institution{University of Georgia}
  \city{Athens}
  \state{GA}
  \country{USA}
}

\author{Md Musfiqur Rahman Sanim}
\email{musfiqur.sanim@uga.edu}
\affiliation{%
  \institution{University of Georgia}
  \city{Athens}
  \state{GA}
  \country{USA}
}

\author{Zhihao Shu}
\email{Zhihao.Shu@uga.edu}
\affiliation{%
  \institution{University of Georgia}
  \city{Athens}
  \state{GA}
  \country{USA}
}

\author{Jiexiong Guan}
\email{jguan@wm.edu}
\affiliation{%
  \institution{William \& Mary}
  \city{Williamsburg}
  \state{VA}
  \country{USA}
}

\author{Xipeng Shen}
\email{xshen5@ncsu.edu}
\affiliation{%
  \institution{North Carolina State University}
  \city{Raleigh}
  \state{NC}
  \country{USA}
}

\author{Miao Yin}
\email{miao.yin@uta.edu}
\affiliation{%
  \institution{University of Texas at Arlington}
  \city{Arlington}
  \state{TX}
  \country{USA}
}

\author{Gagan Agrawal}
\email{gagrawal@uga.edu}
\affiliation{%
  \institution{University of Georgia}
  \city{Athens}
  \state{GA}
  \country{USA}
}

\author{Bin Ren}
\email{bren@wm.edu}
\affiliation{%
  \institution{William \& Mary}
  \city{Williamsburg}
  \state{VA}
  \country{USA}
}
\renewcommand{\shortauthors}{Niu et al.}
\renewcommand{\shorttitle}{\nott{\projectname}}

\begin{abstract}

This work is motivated by recent developments in Deep Neural Networks, particularly the Transformer architectures 
underlying applications such as ChatGPT,  and  the need for performing inference on mobile 
devices. 
\revisioncr{Focusing on emerging transformers (specifically the ones with computationally efficient Swin-like architectures) and large models (e.g., Stable Diffusion and LLMs) based on transformers, 
we observe 
that layout transformations between the computational operators cause a significant slowdown 
in these applications.} This paper presents \projectname, a comprehensive framework 
for eliminating most layout transformations, with the idea that multiple operators 
can use the same tensor layout through careful choice of layout and implementation 
of operations. Our approach is based on classifying 
the operators into four groups, and considering combinations of producer-consumer 
edges between the operators. We  develop a set of methods for searching such layouts. 
Another component of our work is developing efficient memory layouts for 2.5 dimensional memory commonly seen in mobile 
devices.  
\revisioncr{Our experimental results show that \projectname outperforms 5 state-of-the-art DNN execution frameworks on mobile devices across 18 varied neural networks, including CNNs, Transformers with both local and global attention, as well as LLMs. In particular, compared to DNNFusion, \projectname achieves an average speedup of 2.8$\times$, and outperforms TVM and MNN with speedups of 6.9$\times$ and 7.9$\times$, respectively,  on average. }

\end{abstract}
\maketitle

\section{Introduction}  

As Machine Learning (ML), more specifically, Deep Learning (DL) and 
Deep Neural Networks (DNNs) have permeated our every day life,  there is a growing 
need for supporting inference using DL models on the ubiquitous mobile devices.  
From the application side, possibility are endless and go well beyond common 
speech or image recognition. 
On the other hand, we have the growing computational capacity (and continued popularity) 
of smartphones. 

For the past several years, inference with 
even fairly complex models has been feasible on mobile 
devices~\cite{chen2018tvm,Ali-MNN,tensorflow-xla, jia2022codl, shen2021nimble,chen2023speed,niu2021dnnfusion}.
Such inference has the benefit that a user's data does not need to be transmitted to a cloud or a server. 
This, in turn, allows ML models to execute when the device is not connected to the internet, and 
alleviates privacy concerns about sharing personal information that many users frequently have~\cite{manikonda2018s,derner2023safeguards}. 

Recently, Transformers ~\cite{zhao2023survey} have  revolutionized the fields of computer vision (CV) ~\cite{dosovitskiy2020image,bertasius2021space,arnab2021vivit,liu2021Swin,dong2021cswin}
and natural language processing (NLP)~\cite{radford2019language,vaswani2017attention,touvron2023llama,beltagy2020longformer,devlin2018bert,du2022glm}. Transformer-based models uniquely provide long-range dependency handling and global contextual awareness, 
which are driving  existing popular AI applications such as ChatGPT, Bard, and Alexa. 

Studies assessing them as computational workloads~\cite{kim2023full} have identified 
that as compared to the  CNN-based designs, the computation graph representations for transformers 
are more complex, specifically, they have more data flow splits, shuffles, and merges.
One further development has been the emergence and popularity of  (computationally) efficient local-attention 
(Swin-like) ~\cite{liu2021Swin,zhu2023biformer,S3,wang2021crossformer} Transformers that have reduced computational complexity, though at 
the cost of more layout transformations. 

Table~\ref{tab:nn-performance-comparison} summarizes this aspect.
Specifically, three of the older ConvNets (like ResNet50~\cite{he2016deep}), \revisioncr{six newer (Transformer) models, and one decoder-only LLM (Pythia~\cite{biderman2023pythia})}
are compared with respect to the time spent on implicit and explicit data layout transformations (as compared to pure computations).
A majority of the older models spend a relatively 
small fraction of their time on (implicit or explicit) layout transformations. On the other hand,  
Transformer models all consistently spend between \revisioncr{43\% to 70\%} of their time on data transformation. 
Moreover, the execution speed of these models is, on the average, around one order of magnitude 
slower than earlier models. 
It seems likely that increased numbers of data transformation 
\revisioncr{(as indicated in the third column of Table~\ref{tab:nn-performance-comparison})}
cause poor locality during compute-oriented  operations, resulting in a significant slowdown.

\begin{table}
  \scriptsize
  {\setlength{\tabcolsep}{2.7pt}
    \caption{\revisioncr{{\bf Latency and transformation breakdown across various models}. `Lat.' shorts for Latency. `Exp.' refers to the latency associated with explicitly transforming the tensor's layout, such as {\tt Transpose} and {\tt Reshape}. `Imp.' denotes the latency incurred by implicit layout transformations. `Comp.' indicates the latency attributed to the remaining operators. `SD' represents Stable Diffusion model. These results are collected using MNN~\cite{Ali-MNN} on a Snapdragon 8 Gen 2 platform. \revisioncr{MACs means the number of multiply-accumulate operations, and GMACS represents giga MACs per second.} }}
    \label{tab:nn-performance-comparison}
    \begin{tabular}{l|r|r|r|rrr|r}
      \toprule
      \multirow{2}{*}{Model} & \#MACs & \#Layout & Lat. & \multicolumn{3}{c|}{Lat. breakdown (\%)} & Speed \\
      ~                                      & (G)  & transform & (ms)  & Imp. & Exp. & Comp. & (GMACS)    \\
      \midrule
      ResNet50~\cite{he2016deep}             & 4.1  & \revisioncr{3}    & 14    & 4.8  & 0.2  & 95   & {\bf 293}  \\ 
      FST~\cite{johnson2016perceptual}       & 162  & \revisioncr{32}   & 1,506 & 70.7 & 1.8  & 27.5 & {\bf 108}  \\ 
      RegNet~\cite{radosavovic2020designing} & 3.2  & \revisioncr{6}    & 57    & 16.7 & 0    & 83.3 & {\bf 56}   \\
      \midrule
      CrossFormer~\cite{wang2021crossformer} & 5.0  & \revisioncr{208}  & 336   & 15.3 & 55.2 & 29.5 & {\bf 15}   \\
      Swin~\cite{liu2021Swin}                & 4.6  & \revisioncr{242}  & 342   & 14.7 & 54.1 & 31.2 & {\bf 15.2} \\
      AutoFormer~\cite{AutoFormer,S3}        & 4.7  & \revisioncr{233}  & 335   & 13.3 & 54.2 & 32.5 & {\bf 14}   \\
      CSwin~\cite{dong2021cswin}             & 6.9  & \revisioncr{769}  & 703   & 14.3 & 50.2 & 35.5 & {\bf 10}   \\
      \midrule
      \revisioncr{SD-TextEncoder}~\cite{rombach2022high}  & 6.7  & \revisioncr{183} & 133   & 15.1 & 36.3 & 48.6 & {\bf 44}   \\
      \revisioncr{SD-UNet}~\cite{rombach2022high}         & 90   & \revisioncr{533} & 2172  & 19.4 & 42.1 & 38.5 & {\bf 42}   \\
      \revisioncr{Pythia-1B}~\cite{biderman2023pythia}    & 119  & \revisioncr{385} & 3034  & 11.7 & 31.7 & 56.6 & {\bf 39}   \\
      \bottomrule
    \end{tabular}
    \vspace{-1em}
  }
\end{table}

In this work, we take the position that almost all {\em layout transformations} can be eliminated and instead, 
the layout of the tensor that is produced can be chosen to serve various computational operations 
efficiently. This paper presents a systematic framework for enabling elimination 
of such unnecessary memory intensive operations. 
Components of the work include the following: 

\begin{itemize}[leftmargin=*,noitemsep,nolistsep]
    \item Careful study of  the relationship between the computation and input/output data layout of DNN operators, and a high-level operator type classification based on operators' performance sensitivity to input layout and the output layout customizability.
    \item A procedure for layout transformation elimination and an effective heuristic method for selecting the layout for each operator that is not  fused or eliminated.  
    \item A procedure to map the chosen layout to memory hierarchy by taking 2.5D (texture) memory into consideration.
    \item Integration of  the above into a comprehensive framework called \projectname,  which is then implemented on top of a state-of-the-art end-to-end DNN execution framework DNNFusion~\cite{niu2021dnnfusion}. 
\end{itemize}

\revisioncr{\projectname has been extensively evaluated on 18 cutting-edge DNNs, including 4 ConvNet models, 6 Transformer models, and 8 Hybrid  (combining ConvNet and Transformer structures) models on mobile GPUs. 
The evaluation demonstrates a significant speedup compared to 5 state-of-the-art DNN execution frameworks (MNN~\cite{Ali-MNN}, NCNN~\cite{Ni_ncnn_2017}, TFLite~\cite{TensorFlow-Lite}, TVM~\cite{chen2018tvm}, and DNNFusion~\cite{niu2021dnnfusion}).
\projectname reduces the number of \revisioncr{operators} by $21\%$ to $65\%$ compared with other frameworks. In terms of latency, \projectname achieves an average speedup of $2.8\times$ over DNNFusion, a state-of-the-art baseline. Compared with two other popular frameworks, TVM and MNN, \projectname achieves an average speedup of $6.9\times$ and $7.9\times$, respectively.}
Furthermore, \projectname enhances cache utilization and reduces memory pressure, enabling the execution of some models on resource-constrained devices while other frameworks may encounter challenges.

\section{Background and Motivation}

\subsection{DNN Recent Advances: Transformers. }

The Transformer architecture has become the dominant paradigm in deep learning space, 
leveraging {\em attention mechanism}  to focus on different parts of the input sequence while 
generating each part of the output sequence~\cite{vaswani2017attention,radford2019language,beltagy2020longformer,devlin2018bert}. 
One notable challenge with the standard (or global) attention mechanism is its computational and space complexities, both of which are $O(n^2)$, where $n$ represents the length of the input sequence. 
 DL practitioners have built upon the foundational (vanilla)  Transformer model by introducing local-attention Transformer 
models~\cite{ding2020lanet,liu2021Swin,
lin2023scaleaware,
AutoFormer,S3} that reduce computational complexity. 

\begin{figure}[t]
  \centering
  \includegraphics[width=0.75\columnwidth]{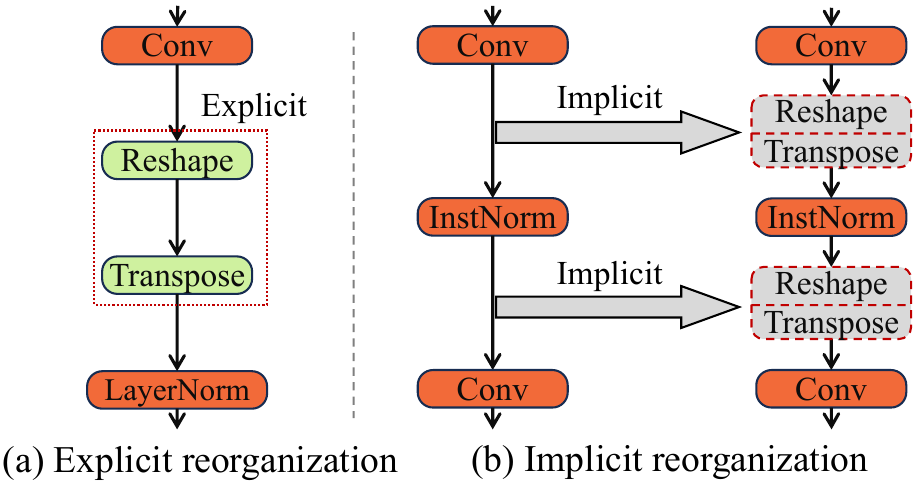}  
  \caption{{\bf Examples of layout transformation in DNNs.}}
  \label{fig:challenge_relayout_types}
  \vspace{-1em}
\end{figure}

Local attention focuses on a subset of input tokens (typically within a window) at a time and requires less computation. 
To achieve computational reduction,  frequent explicit shape/data reorganization (Reshaping,
Transposing, Gathering) is used to split and transpose segments within
the input data into these smaller windows.
Another trend involves combining traditional ConvNets with Transformers and designing new model 
structures~\cite{dong2021cswin,zhu2023biformer,cai2022efficientvit, han2023flatten,xu2021co,dai2021coatnet,wang2020axial} to benefit from both structures.
These new structures introduce implicit data reorganization due to different layout preferences for various operators.
Figure ~\ref{fig:challenge_relayout_types} shows the two types of data reorganization, which we will discuss 
next.  

\subsection{Motivation and Research Issues} 
\label{subsec:motiv}

To motivate the key optimizations we are proposing, we further discuss  two examples from Figure~\ref{fig:challenge_relayout_types}. 
In sub-figure (a), we show  two computational operations, a {\tt Conv} (convolution) and a {\tt LayerNorm}. 
In between these two, the programmer explicitly inserts two operations, a {\tt Reshape} and a {\tt Transpose}. 
A reshape operation changes the number and sizes of dimensions for the tensor -- for example, a $5 \times 5 \times 5 $  
3-D tensor can be recast as a $5 \times 25$ 2-D matrix. A subsequent transpose operation can next convert it to a $25 \times 5$ matrix.

In Figure~\ref{fig:challenge_relayout_types} (a), the {\tt Reshape} and {\tt Transpose} are considered {\em explicit}, 
in the sense they are inserted explicitly by a model implementer while working with a framework such as MNN~\cite{Ali-MNN}. 
In Figure~\ref{fig:challenge_relayout_types} (b) we  show what we consider as an {\em implicit}  
layout transformation. In this example, a {\tt Conv} operation is followed by an {\tt InstNorm} or Instance Normalization 
operation, which is then followed  by another {\tt Conv}. A framework such as MNN inserts both {\tt Reshape} and 
{\tt Transpose} examples before and after the {\tt InstNorm}, aligning them with predefined input layouts.

The underlying reason behind these operations is that different operators in a DNN might require input tensors with different shapes and/or layouts. For instance, fully connected operators require a {\em flattened input}, whereas convolutional operators require multi-dimensional inputs to perform spatial operations. 
A transpose operation consumes high memory bandwidth, and furthermore, can reduce locality for the operations that follow 
the transpose. Eliminating reshape and transpose operations, and performing 
the two operations, i.e., those before and after the transpose, on the same layout  should be able to improve performance.  However, eliminating such 
layout transformation involves many important (inter-dependent) questions. 

\begin{table}
  \scriptsize
  {\setlength{\tabcolsep}{2.3pt}
    \caption{{\bf Memory comparison on mobile GPUs.}}
    \label{tab:gpu-memory-comparison}
    \begin{tabular}{ll|rr}
      \toprule
      \multicolumn{2}{c|}{Characteristics} & 1D Buffer & {\bf 2.5D Texture} \\
      \midrule
      \multirow{2}{*}{Computation} & Acceleration engine       & N & {\bf Y}           \\
      ~                            & Automatic bounds checking & N & {\bf Y}           \\
      ~                            & Hardware interpolation    & N & {\bf Y}           \\
      \midrule
      \multirow{5}{*}{Data access} & Organization & Contiguous & {\bf Multidimensional}    \\
      ~                            & Addressing & Pointer-based & {\bf Coordinates} \\
      ~                            & Dedicated cache & No & {\bf Yes}    \\
      ~                            & Data locality & 1D & {\bf 2.5D}  \\
      ~                            & Direct memory access on CPU & Yes & {\bf No}   \\
      \bottomrule
    \end{tabular}
  }
\end{table}

\begin{figure}[t]
  \centering
  \includegraphics[width=0.4\textwidth]{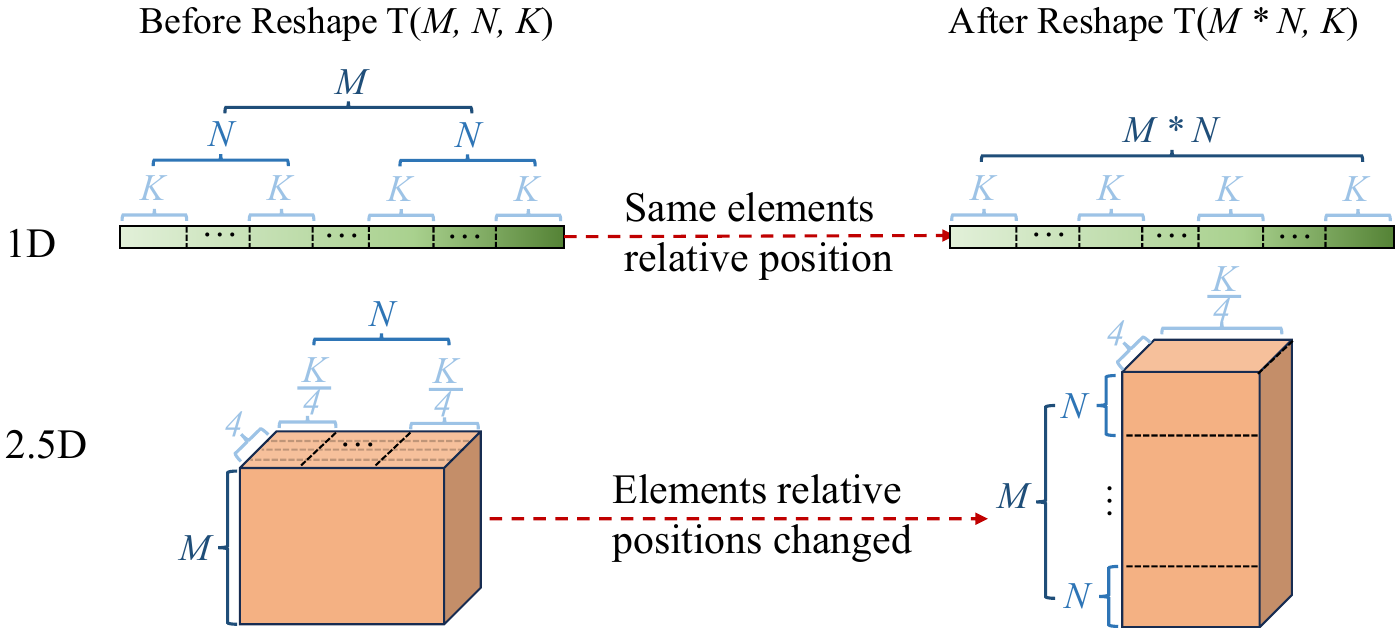}
  \caption{{\bf Layout transformation and 2.5D memory.}}
  \label{fig:challenge_relayout_matters}
\end{figure}

\begin{itemize}[leftmargin=*,noitemsep,nolistsep]

\item {\bf Q1:} In a large and complex computational  graph capturing a large DNN, deciding when two consecutive operations should use the same layout of data. 

\item {\bf Q2:} For two operations where we decide to use the same layout, choosing the layout that leads to efficient execution 
of both operands. 

\item {\bf Q3:} Implementing an operation efficiently for a chosen layout (distinct from the original layout), including deciding access 
pattern and simplifying index computations. 

\item {\bf Q4:} Mapping the chosen layout to memory hierarchy, especially the 2.5D memory (further discussed 
in Section~\ref{subsec:texture}). 

\end{itemize} 

\noindent Many of these issues have been addressed previously in computer systems and compiler community, though not specific to our 
target workload. The closest work in this area is on integrating data layout selection and loop 
transformations~\cite{manjikian1997fusion,kandemir1998improving,o2002integrating,shirako2020affine,smith2021pure,feng2023tensorir}. However, 
the most significant difference in our work is that prior efforts, motivated by traditional scientific workloads, 
have focused on one nested loop. In comparison, with transformers (or even other DNNs), the challenge is making 
this selection among a Directed Acyclic Graph (DAG) of operators (i.e. the computational graph). Other differences 
in the work involve deciding when eliminating a transpose is beneficial or not, implementing multiple 
operators efficiently on the same layout, and physically mapping to 2.5D memory.  

\subsection{GPU texture memory} 
\label{subsec:texture}

The GPU texture memory specializes in improving two-dimensional spatial locality.
Since each cache element can be a vector with a fixed length of four, and the cache itself has two dimensions (referred to as width and height), we refer it to as 2.5D memory.
Originally designed to facilitate graphical rendering processes, this design also offers significant advantages for stencil and similar computations.
For instance, using 2.5D texture memory for convolution operations can result in a 3.5$\times$ reduction in latency compared to 1D buffer memory \cite{jia2022codl}.
Table~\ref{tab:gpu-memory-comparison} summarizes the main differences between GPU texture memory and 1D buffer memory.

Coupled with the advantages associated with spatial locality, 
there are also some challenges. 
Figure ~\ref{fig:challenge_relayout_matters} shows an example of {\tt Reshape} in 1D and 2.5D memory.
As a background, reshaping a tensor involves changing its shape without altering the underlying data order.
For a 1D memory,  due to the linear format, 
reshaping simply implies interpreting the same data with a different number of dimensions. 
\revisioncr{The overhead is negligible as it only involves changing the metadata of the tensor.}
However, for 2.5D memory,  because spatial relationships between data points are important (consider image processing or matrix computations),  
reshaping a tensor in 2.5D memory involves more complex processes of reordering the data layout while maintaining inherent relationships within the data.
\revisioncr{Furthermore, due to the limited bandwidth, the data transformation overhead is crucial when an operation requires moving both metadata and actual data (such as explicit and implicit data transpose) in 1D linear buffer and 2.5D texture memory on mobile GPUs.}

\section{Design of SmartMem}

Our framework has three components, which address the four questions listed 
earlier in Section~\ref{subsec:motiv}: 1) an operator type classification based on operators' performance sensitivity to input layout as well as the output layout customizability, and an analysis built on this operator type classification (to answer Q1), 2) a layout transformation elimination procedure and a method for selecting 
the optimal layout for each (not eliminated)  operator based on the operator type classification and layout transformation analysis (to answer Q2 and Q3), and 3) a further optimization procedure that maps the chosen layout to memory hierarchy by taking 2.5D memory into account (to answer Q4).

\begin{table}[!t]
  \scriptsize {
    \setlength{\tabcolsep}{2.5pt}
    \renewcommand{\arraystretch}{1.2}
    \caption{{\bf Operator classification based on input layout dependency and output layout flexibility.} `Comp.' stands for Computation.}
    \label{tab:design-operator-classification}
    \begin{tabular}{l|r|r}
      \toprule
      \diagbox[width=1.8cm]{\makecell{Comp. \\performance}}{\makecell{Output \\layout}}  & \makecell{Variable (Computation \\order dependent)}  & \makecell{Fixed} \\
      \midrule
      \multirow{4}{*}{\makecell{Input layout \\dependent (ILD)}}
      ~ & \opname{Conv}$C_{\tdim{4}}^{l} \leftarrow A_{\tdim{4}}^{l_1} \ast B_{\tdim{4}}^{l_2}$ & \opname{Reshape}$B_{\tdim{n}}^{L} \leftarrow A_{\tdim{m}}^{l_1}$       \\
      ~ & \opname{MatMul}$C_{\tdim{2}}^{l} \leftarrow A_{\tdim{2}}^{l_1} \cdot B_{\tdim{2}}^{l_2}$ & \opname{Transpose}$B_{\tdim{n}}^{L} \leftarrow A_{\tdim{n}}^{l_1}$  \\
      ~ & \opname{LayerNorm}$C_{\tdim{n}}^{l} \leftarrow A_{\tdim{n}}^{l_1} \odot B_{\tdim{2}}^{l_2}$ & \opname{DtoS}$B_{\tdim{n}}^{L} \leftarrow A_{\tdim{n}}^{l_1}$ \\
      ~ & \opname{Softmax}$C_{\tdim{n}}^{l} \leftarrow A_{\tdim{n}}^{l_1} \odot B_{\tdim{2}}^{l_2}$ & \opname{StoD}$B_{\tdim{n}}^{L} \leftarrow A_{\tdim{n}}^{l_1}$  \\
      \midrule
      \multirow{2}{*}{\makecell{Input layout \\ independent (ILI)}} 
      ~ & \opname{Unary}$B_{\tdim{n}}^{l} \leftarrow A_{\tdim{n}}^{l_1}$ & \opname{Gather}$B_{\tdim{n}}^{L} \leftarrow A_{\tdim{m}}^{l_1}$  \\
      ~ & \opname{Add}$C_{\tdim{n}}^{l} \leftarrow A_{\tdim{n}}^{l_1} + B_{\tdim{n}}^{l_1}$ & \opname{Slice}$B_{\tdim{n}}^{L} \leftarrow A_{\tdim{n}}^{l_1}$ \\
      \bottomrule
      \multicolumn{3}{l}{Unary refers to an operator applying a function to each element of a single input.} \\
      \multicolumn{3}{l}{DtoS and StoD mean DepthToSpace and SpaceToDepth, respectively.} \\
    \end{tabular}
  }
\end{table}

\subsection{Operator Classification and Analysis}

The foundation of our method is a novel classification system for operators commonly seen in 
DNNs. 
Any given operator in our target workload is classified along two dimensions. The first dimension 
is whether the performance of the computation depends upon the input layout or is independent.  
The second dimension is whether the output layout is customizable (perhaps in view of the 
computation pattern chosen for operator's implementation). These two dimensions result in four quadrants and each 
operator can be mapped to one quadrant. In some cases, one operator may be placed in 
different quadrants depending on whether the layout of its different operands is the same 
or different.  Table~\ref{tab:design-operator-classification} shows sample operators for each quadrant.

To explain these quadrants, we start at the bottom left and consider the addition  operation: 
\[ C_{\tdim{n}}^{l} \leftarrow A_{\tdim{n}}^{l_1} + B_{\tdim{n}}^{l_1} \]  

\begin{table}[!t]
  \scriptsize {
    \setlength{\tabcolsep}{4pt}
    \renewcommand{\arraystretch}{1}
    \caption{{\bf Operator type definition.}}
    \label{tab:design-operator-definition}
    \begin{tabular}{c|c}
      \toprule
      Name  & Definition \\
      \midrule
      Input Layout Dependent & \multirow{2}{*}{$B_{\sigma(i),\sigma(j),\ldots,\sigma(n)} = \mathcal{O}^{ILD-Variable}_{\pi(i),\pi(j),\ldots,\pi(m)}(A_{i,j,\ldots,m})$} \\
       \& Variable & ~ \\\hline
      Input Layout Dependent & \multirow{2}{*}{$B_{i,j,\ldots,n} = \mathcal{O}^{ILD-Fixed}_{\pi(i),\pi(j),\ldots,\pi(m)}(A_{i,j,\ldots,m})$} \\
       \& Fixed & ~ \\\hline
      Input Layout Independent & \multirow{2}{*}{$B_{\sigma(i),\sigma(j),\ldots,\sigma(n)} = \mathcal{O}^{ILI-Variable}(A_{i,j,\ldots,m})$} \\
       \& Variable & ~ \\\hline
      Input Layout Independent & \multirow{2}{*}{$B_{i,j,\ldots,n} = \mathcal{O}^{ILI-Fixed}(A_{i,j,\ldots,m})$} \\
       \& Fixed & ~ \\\bottomrule
    \end{tabular}
  }
\end{table} 

In the above operation, tensors $A$ and $B$ have the same layout denoted as $l_1$. Having an 
identical layout, and with an addition operation needing to touch each element once, the computational 
performance is not sensitive to the layout $l_1$, i.e., addition can be performed efficiently with 
traversal of the two matrices matching their (identical) physical layout.  
At the same time, the layout of the output tensor $C$ can be customized, for example, based 
on the needs of the downstream operations.  

Moving clockwise, we consider matrix multiplication example $C_{\tdim{2}}^{l} \leftarrow A_{\tdim{2}}^{l_1} \cdot B_{\tdim{2}}^{l_2}$. 
As a multiplication operation involves temporal reuse of data, the performance of the operation 
is clearly sensitive to the input layouts (even if the layouts $l_1$ and $l_2$ are identical). At 
the same time, the output layout can be customized for this application.  
An operator like {\tt Transpose}, on the other hand, has a well-defined output layout (dependent 
upon the input layout). Given the memory transformations involved, the performance can be 
sensitive to the layout of the input operand. 
Finally, consider an operator like {\tt Slice}. Because of simple selection  involved, 
the performance is insensitive to the layout of the input. Moreover, by definition, 
the output layout has to match the input layout and is therefore not customizable 

We next formally define these four quadrants (or types) of operators so that we can classify and place a given 
operator into this table.
Table ~\ref{tab:design-operator-definition} shows the formal definition to these four quadrants (or types) of operators 
so that we can classify and place a given 
operator into Table~\ref{tab:design-operator-classification}.

Take Input Layout Dependent \& Variable (ILD-Variable) as an example.
Let  \( A \) and \( B \) represent the input tensor and the output tensor, respectively, and \( \mathcal{O}^{ILD-Variable} \) denotes the operator. The permutations \( \pi \) and \( \sigma \) represent the flexibility in processing and generating  data, respectively.
  More specifically,  The permutation \( \pi \) represents a rearrangement of these dimensions that affects the input layout, and \( \sigma \) represents a rearrangement of the computed result's dimensions affecting the output layout. 
   Denoting mathematically, 

    \[ B_{\sigma(i),\sigma(j),\ldots,\sigma(n)} = \mathcal{O}^{ILD-Variable}_{\pi(i),\pi(j),\ldots,\pi(m)}(A_{i,j,\ldots,m}) \]
    
\noindent The input tensor is accessed as  $  A'_{\pi(i),\pi(j),\ldots,\pi(m)} = A_{i,j,\ldots,m}$ during the computation, 
    where \( A' \) is the tensor \( A \) with its layout altered according to \( \pi \). 
    This rearrangement can lead to different memory access patterns, which might impact cache performance and processing speed. 
    The computation is then performed, and the result is organized in the output tensor \( B \) as $ B'_{i,j,\ldots,n} = \mathcal{O}(A'_{\pi(i),\pi(j),\ldots,\pi(m)})$. Finally, the output tensor \( B \) is produced with a layout determined by \( \sigma \), i.e. $B_{\sigma(i),\sigma(j),\ldots,\sigma(n)} = B'_{i,j,\ldots,n}$.

Each of these categories reflects a different combination of data layout considerations and computational patterns, which are crucial for optimizing the performance of DNNs. 
   The permutation functions \( \pi \) and \( \sigma \) accommodate the flexibility in data layout and output generation, which can be exploited for performance enhancements such as minimizing memory traffic, improving cache usage, or optimizing parallel execution strategies.

\begin{table}[!t]
  \scriptsize {
    \setlength{\tabcolsep}{1.5pt}
    \renewcommand{\arraystretch}{1.2}
    \setlength\aboverulesep{0pt}
    \setlength\belowrulesep{0pt} 
    \caption{{\bf Operator combination - action.} ILD \& Var is short for ILD \& Variable.}
    \label{tab:design-operator-combination-action}
    \begin{tabular}{l|l|r|r|r|r}
      \toprule
      \multicolumn{2}{l|}{\multirow{2}{*}{ \diagbox[width=2.0cm]{\bf First}{\bf Second} }}
                             & \multirow{2}{*}{\makecell{ILD \& Variable}}   & \multirow{2}{*}{\makecell{ILI \& Variable}} & \multirow{2}{*}{\makecell{ILD \& Fixed}} & \multirow{2}{*}{\makecell{ILI \& Fixed}} \\ 
      \multicolumn{2}{l|}{~} & ~ & ~ & ~ & ~ \\\midrule
      ILD \& Var & Action & \cellcolor{cell1}Keep both     & \cellcolor{cell2}Try fuse      & \cellcolor{cell3}Eliminate 2nd  & \cellcolor{cell3}Eliminate 2nd \\\hline
      
      ILI \& Var & Action & \cellcolor{cell2}Try fuse      & \cellcolor{cell2}Try fuse      & \cellcolor{cell3}Eliminate 2nd  & \cellcolor{cell3}Eliminate 2nd \\\hline
      
      ILD \& Fixed & Action & \cellcolor{cell4}Eliminate 1st & \cellcolor{cell4}Eliminate 1st & \cellcolor{cell5}Eliminate both & \cellcolor{cell5}Eliminate both \\\hline
      
      ILI \& Fixed & Action & \cellcolor{cell4}Eliminate 1st & \cellcolor{cell4}Eliminate 1st & \cellcolor{cell5}Eliminate both & \cellcolor{cell5}Eliminate both \\\bottomrule
    \end{tabular}
  }
\end{table}

\subsection{Layout Transformation Elimination Analysis}

The aforementioned operator classification reveals the relationship between the computation and input/output layouts of an operator.  
As stated above, the four combinations are: 
input layout dependent and variable output (ILD \& Variable), input layout independent and variable output (ILI \& Variable), input layout dependent and fixed output (ILD \& Fixed), and input layout independent and fixed output (ILI \& Fixed). From a performance optimization perspective, we observe that their  ``optimization complexity'' gradually decreases. For example, ILD \& Variable requires us to be aware of both the input and output layouts while ILI \& Fixed has no requirement about either of the  layouts. 

\begin{table}[!t]
  \scriptsize {
    \setlength{\tabcolsep}{2.2pt}
    \renewcommand{\arraystretch}{1.2}
    \setlength\aboverulesep{0pt}
    \setlength\belowrulesep{0pt}
    \caption{{\bf Operator combination and their corresponding design decisions.}}
    \label{tab:design-operator-combination-outcome}
    \begin{tabular}{l|l|r|r|r|r}
      \toprule
      \multicolumn{2}{l|}{\multirow{2}{*}{ \diagbox[width=1.5cm]{\bf First}{\bf Second}}}
                             & \multirow{2}{*}{ILD \& Var}   & \multirow{2}{*}{ILI \& Var} & \multirow{2}{*}{ILD \& Fixed} & \multirow{2}{*}{ILI \& Fixed} \\ 
      \multicolumn{2}{l|}{~}     & ~ & ~ & ~ & ~ \\\midrule
      ILD \&   & Output & \cellcolor{cell1}ILD \& Variable$^*$ & \cellcolor{cell2}ILD \& Variable & \cellcolor{cell3}ILD \& Variable & \cellcolor{cell3}ILD \& Variable  \\
      Var & Layout & \cellcolor{cell1}Search both         & \cellcolor{cell2}Search fused    & \cellcolor{cell3}Search 1st      & \cellcolor{cell3}Search 1st  \\ \hline
      
      ILI \&   & Output & \cellcolor{cell2}ILD \& Variable & \cellcolor{cell3}ILI \& Variable & \cellcolor{cell3}ILI \& Variable & \cellcolor{cell3}ILI \& Variable  \\
      Var & Layout & \cellcolor{cell2}Search fused    & \cellcolor{cell3}No search       & \cellcolor{cell3}No search       & \cellcolor{cell3}No search   \\\hline
      
      ILD \&   & Output & \cellcolor{cell4}ILD \& Variable & \cellcolor{cell4}ILI \& Variable & \cellcolor{cell5}N/A & \cellcolor{cell5}N/A     \\
      Fixed    & Layout & \cellcolor{cell4}Search 2nd      & \cellcolor{cell4}No search       & \cellcolor{cell5}No search    & \cellcolor{cell5}No search   \\\hline
      
      ILI \&   & Output & \cellcolor{cell4}ILD \& Variable & \cellcolor{cell4}ILI \& Variable & \cellcolor{cell5}N/A & \cellcolor{cell5}N/A     \\
      Fixed    & Layout & \cellcolor{cell4}Search 2nd      & \cellcolor{cell4}No search       & \cellcolor{cell5}No search    & \cellcolor{cell5}No search   \\\bottomrule
      \multicolumn{6}{l}{$^*$ Both operators are ILD \& Variable.} \\
    \end{tabular}
  }
\end{table}

Based on this key insight, Table~\ref{tab:design-operator-combination-action} summarizes \projectname's computation optimizations (on a DNN computational graph) 
for each pair of DNN operators. Specifically, \projectname has four levels of computation optimizations (marked with different colors): keeping both operators, trying to fuse them, eliminating one operator (either the first or the second), and eliminating both operators, which represent the optimization opportunities from low to high. 
Correspondingly, Table~\ref{tab:design-operator-combination-outcome} summarizes the resulting output type and the input/output layout search policies after the computation optimizations explained above. The resulting output type is decided by the operator with a higher optimization complexity or the preserved operator, for example, after the computation optimization of a pair of operators with ILD \& Variable and ILI \& Variable types, respectively, the resulting (fused) operator is ILD \& Variable, and after eliminating the second operator in a pair of ILD \& Variable and ILD \& Fixed operators, the resulting operator is in ILD \& Variable, too.   

The input/output layout search also has four levels (marked with different colors): searching input and output layouts for both operators, searching input and output layouts for the fused operator, searching for either the first or the second, and no need to perform any search operation, representing the varied levels of optimization processing difficulties from high to low. It is worth noting that the layout search only happens for the operator pairs involving ILD \& Variable.
To explain the ideas, consider a pair of operators,  i.e., {\tt Conv+Reshape} (ILD \& Variable + ILD \& Fixed) as an example. Table~\ref{tab:design-operator-combination-action}  implies that {\tt Reshape} can be eliminated\footnote{The subtle difference between operator fusion and elimination will be elaborated in Section~\ref{sec:layout-trans-eliminate-eliminate}.}. According to Table~\ref{tab:design-operator-combination-outcome}, the preserved operator ({\tt Conv}) is still in ILD \& Variable and \projectname needs to search for its input layout.

To achieve both computation and layout selection optimizations, \projectname specifically answers these three questions:
\begin{itemize}
[leftmargin=*,noitemsep,nolistsep]
    \item How to fuse operators? Specifically how to decide if an operator fusion is legal and profitable?
    \item How to correctly and effectively eliminate any operators (specifically, the layout transformation operators in the types of ILD \& Fixed or ILI \& Fixed)? 
    \item How to select data layout for fused/preserved operators?  
\end{itemize}

With respect to the first question,  \projectname relies on the techniques based on 
the DNNFusion project~\cite{niu2021dnnfusion} to decide if an operator fusion is legal. Based on DNNFusion, the subsequent operator elimination and layout selection designed in \projectname bring more opportunities for beneficial fusions,  enhancing  execution performance  
(as we show through our evaluation results in Section~\ref{sec:eval}). Thus, this section mainly focuses on the new operator elimination and layout selection techniques to answer the last two questions.

\subsubsection{Operator Elimination based on Index Comprehension}\label{sec:layout-trans-eliminate-eliminate}

All cases (except the first marked by red) in Table~\ref{tab:design-operator-combination-action} can 
be optimized by an operator fusion (by following the rules in DNNFusion~\cite{niu2021dnnfusion}). Going beyond operator 
fusion, it turns out that a more advanced optimization called {\em Operator Elimination} can be leveraged for cases involving any operator with a {\em Fixed} output type (i.e., the cases marked by either yellow or green) to further improve the performance of the memory access in the fused operator. More specifically, after fusing a sequence of layout transformation operators, 
these operators can be replaced by index computations for 
the operator it is fused with. 

\begin{figure}[!t]
  \centering
  \includegraphics[width=0.85\columnwidth]{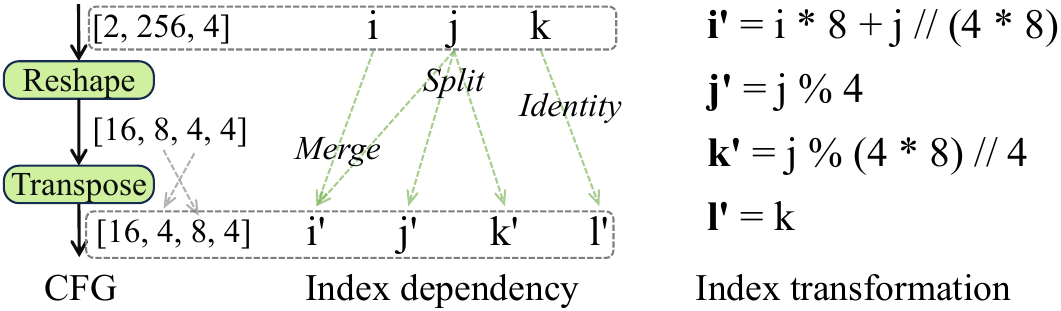}
  \caption{{\bf An example of index dependency and transformation.} left:   a computational graph comprising {\tt Reshape} and {\tt Transpose},  middle:   index dependency and transformation from the input, right: index computation (no opt.).}
  \label{fig:example_index_dependency}
\end{figure} 

\noindent\underline{\em Strength reduction on index computation.}
To further reduce the overhead  for  index computation 
during  data loading and storage, we  propose the following optimizations for index computation.
As a background, {\tt Reshape} transforms an input with shape $[d_1, \dots, d_{m}]$ into an output with shape $d'_1, \dots, d'_{n}$ where the product of the new shape must equal to the old shape.
{\tt Transpose} permutes the dimensions according to a given order -- formally: $Out_{i, \dots, i_{n}} \leftarrow In_{\pi(i), \dots,\pi(i_{m})}$.
However, it turns out that using the linear representation for all indexes directly leads to redundant computations, especially when multiple {\tt Reshape} and {\tt Transpose} operations are stacked together. 
Instead, our strength reduction strategy analyzes the index dependencies between consecutive layers. 
As shown in Figure \ref{fig:example_index_dependency}, we define the index dependencies as {\em identity}, {\em split}, and {\em merge} based on static shape information in the computational graph. 
This allows us to transform indexes according to their dependency types using operands such as ``$\sslash, \%$'' (used in {\em split}) and ``$*, +$'' (used in {\em merge}), ``='' (used in {\em identity}). 
Since modular and divide operations are expensive on GPUs, we also simplify these terms by applying mathematical strength reduction rules. 
For example, if $i$, ($C_a$, $C_b$) are a variable index and constants respectively, then $ i \% C_a \% C_b $ can be reduced to  $i \% C_b$ when $C_a \% C_b \equiv 0$. This reduction commonly occurs when there are layout variations involved in index computation.

An additional point to be mentioned is that before  index computation simplification, the fused operator is 
combining the logic of multiple layout transformations, rendering memory accesses fragmented and difficult to optimize.  After the simplification process described earlier, 
the memory access pattern is more straightforward and exposes 
aggressive optimization opportunities.

\subsubsection{A Reduction Dimension Based Layout Selection} 
At a high-level, we have the problem of selecting layout for all tensors throughout the computational 
graph. 
This  global layout selection involves a large search space~\cite{liu2019optimizing} and can be 
considered NP-hard, as evidenced by the connection to the Partitioned Boolean Quadratic Problem (PBQP)~\cite{anderson2018optimal}. To keep the process at manageable costs, we develop a new heuristic solution based on {\em Reduction Dimension} that comprises two main steps.  First, we  conduct  a local layout selection  
for tensors associated with individual edges in the computational graph (i.e., for pairs of operators), in which, the source  operator is  the producer while the sink  is a consumer of a tensor.  It is worth noting  that we only need to handle the edges/operator pairs involving ILD\&Variable (where we have denoted ``search layout''  in Tables~\ref{tab:design-operator-combination-action} and ~\ref{tab:design-operator-combination-outcome}).  
However, this process needs to be augmented through an additional step when we need to  find the layout for producers with multiple consumers.
Both steps rely on the knowledge of the {\em reduction dimension}, which we discuss next.

\noindent\underline{\em Our heuristic: reduction dimension.}
Reduction dimension(s) for an operand of an operator is  the (set of) dimension along which  data  elements are involved 
in an aggregation. 
Take {\tt MatMul} with \( A_{i, k} \) and \( B_{k, j} \) as inputs as an example. Its reduction dimension is \( k \) for both matrices $A$ and $B$.

To examine how the notion of reduction dimensions is used, we revisit 
 Tables~\ref{tab:design-operator-combination-action} and ~\ref{tab:design-operator-combination-outcome}. 
 After multiple rounds of operator fusion and operator elimination, all preserved operators are ILD \& Variable  -- 
 this is because all operators in other types including ILI \& Variable are fused into ILD \& Variable eventually. 
The next step  is to find the layout on each edge/operator pair (with ILD \& Variable type). Specifically, this step forces the first operator of each edge (i.e., producer) to generate the data layout preferred by the second operator (consumer). 
\begin{figure}[!t]
  \centering
  \includegraphics[width=0.9\columnwidth]{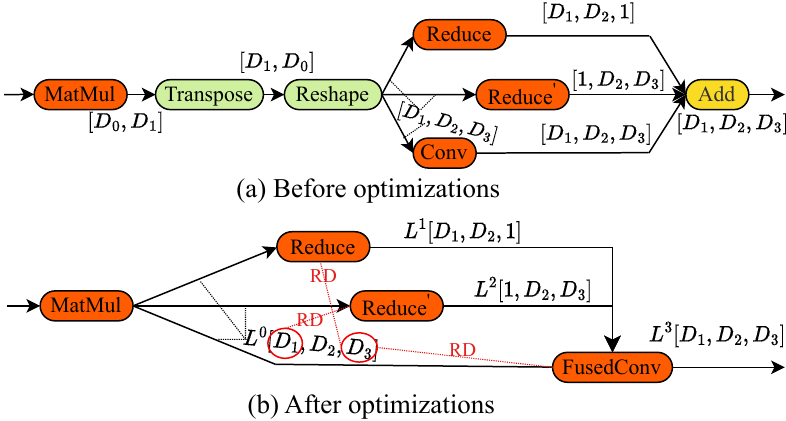}
  \caption{{\bf Examples of reduction dimensions and layout selection.} $D_0, D_1, D_2, D_3$ represent dimensions in intermediate results. RD shorts for reduction dimension, and $L$ means the memory layout. {\tt Add} broadcasts its input shapes to match the shape of the largest one.}
  \label{fig:example_reduction_dimension_cfg}
  \vspace{-0.1cm}
\end{figure} 
In our approach, this preferred data layout is decided by the reduction dimension of the second operator. For example, for the above {\tt MatMul} example, we should keep the elements in both $A_{i, k}$ and $B_{k, j}$ along the reduction dimension ($k$) continuously stored in the memory. 
The insight here is that forcing the producer to  generate  a layout based on the reduction dimension of the consumer
incurs relatively low added overhead as compared to other options. Specifically, sub-optimally 
writing results turns out to be better than sub-optimally reading input data. 
\revisioncr{Microbenchmarking  experiments 
comparing two versions (a.  version that optimizes read performance, 
and b. version that optimizes write performance)
using three operators ({\tt Conv}, {\tt MatMul}, and {\tt Activation}) show  speedups  of  $1.7\times$, $1.4\times$, and $1.1\times$, respectively 
for version a.} 
\projectname also includes a set of optimized tensor (matrix) layouts that are designed for both 1D memory and 2.5D memory for the producer to select. Section~\ref{sec:map-texture-other-opt} elaborates these sample optimized layouts for 2.5D  memory.

\noindent\underline{\em Global: layout selections based on reduction dimension.}
Sear-\\ching for optimal layouts when one producer operator may have multiple consumer operators becomes challenging. 
We optimize  the layout for the producer based on the collective needs from its consumers.
Specifically,  we optimize corresponding to the first $k$ reduction dimensions  required by the consumers, where $k$ is the 
number of dimensions along which we can perform continuous memory access without any linearization and index computation. For example, $k = 2$ for 2.5D  memory. 
Section~\ref{sec:map-texture-other-opt} will elaborate more details for 2.5D memory. If consumers require more than $k$ optimized layouts, \projectname needs to maintain several copies of data with different layouts, and each layout is in this optimized combined format.

\noindent\underline{\em Example.} Figure ~\ref{fig:example_reduction_dimension_cfg} illustrates our reduction dimension-based layout selection approach on a simplified computation graph.
Figure ~\ref{fig:example_reduction_dimension_cfg}(a) shows the original computation graph with a series of operators and the dimensions of their intermediate results. 
{\tt Transpose} and {\tt Reshape} (which  splits the dimension $D_0$ into $D_2$ and $D_3$ for all successor operators) 
are inserted here for  aligning the dimension for different kinds of operators ({\tt MatMul} and {\tt Conv}). 
Figure ~\ref{fig:example_reduction_dimension_cfg} (b) demonstrates an optimized computation graph with our reduction dimension-based layout selection. 
 {\tt Transpose} and {\tt Reshape} are both ILD-Fixed and hence can be eliminated (as shown in Table ~\ref{tab:design-operator-combination-action}).
Then the output tensor of {\tt MatMul} is consumed directly by {\tt Reduce}, {\tt Reduce}$'$, and {\tt  Conv(FusedConv)};
however, these operators suggest two reduction dimensions 
in which, {\tt Reduce}$'$ has a reduction dimension of $D1$ while {\tt Reduce} and {\tt  Conv(FusedConv)} have a common reduction dimension of $D3$.
Assuming  a mobile GPU with 2.5D memory (and cache), \projectname can combine these two reduction dimension requirements (i.e., optimized layout requirements) for consumers of {\tt MatMul} into an uniform data layout ($L^0$ as shown in the figure) and preferably map $D1$ and $D3$ to the two memory dimensions that allow continuous and direct indexed memory access.  The layout generations of {\tt Reduce} ($L^1$) and {\tt Reduce}$'$ ($L^2$) are more straightforward because they have no reduction dimension requirement. 
We will explain more details in Section~\ref{sec:map-texture-other-opt}.

\begin{figure}[!t]
  \centering
  \includegraphics[width=1\columnwidth]{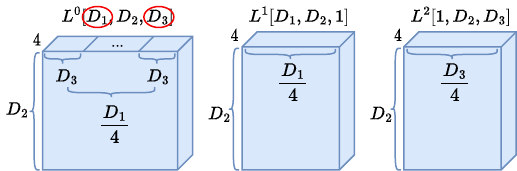}
  \caption{{\bf Sample layouts on 2.5D memory.} Red circles denote reduction dimensions.}
  \label{fig:example_cd_layout}
\end{figure}

\subsection{Mapping Tensor to Texture Memory and Other Optimizations}\label{sec:map-texture-other-opt}

2.5D memory (and corresponding  dedicated read-only cache) on mobile GPUs allows more flexible index computation and eliminates index linearization if a tensor's dimension is less than 3. It also facilitates better exploring data reuse opportunities for 2D and 3D tensors. Thus, \projectname also leverages 2.5D memory to further improve our optimized tensor layout design and memory access as mentioned 
earlier. 

\noindent\underline{\em Optimized tensor layout on 2.5D memory.}
Figure~\ref{fig:example_cd_layout} shows 3 sample tensor layouts when mapping a 3D tensor with varied shapes to 2.5D memory, corresponding to $L_0$, $L_1$, and $L_2$ in Figure~\ref{fig:example_reduction_dimension_cfg}, respectively ($L_3$ depends on its consumers). $D1$, $D2$, and $D3$ denote the dimensions of these tensors -- both $L^1$ and $L^2$ have a dimension of size 1. Specifically, the red circles denote that these dimensions are specified as reduction dimensions by the consumer operators of this tensor, for example, $L^0$ has two reduction dimensions, $D1$ and $D3$ decided by {\tt Reduce'} and {\tt FusedConv} (in Figure~\ref{fig:example_reduction_dimension_cfg}), respectively, while $L^1$ and $L^2$ have no reduction dimensions.

It  is beneficial to store the data along a reduction dimension continuously to improve data locality and allow better SIMD load and reduction operations. Thus, to map a tensor with two reduction dimensions (e.g., $L^0$ with $D1$ and $D3$) to 2.5D memory, \projectname partitions one reduction dimension ($D1$). Each such  partition has  $k$ elements ($k = 4$ in this example to 
match the size of the 0.5D in  2.5D) and stores $k\times$ <another reduction dimension>  elements as a chunk along the  dimensions of 2.5D memory that can be accessed in both directions. This layout results in efficient memory access and SIMD operations for consumer operators using either $D1$ or $D3$ as their reduction dimensions.

\begin{figure}[!t]
  \centering
  \includegraphics[width=0.95\columnwidth]{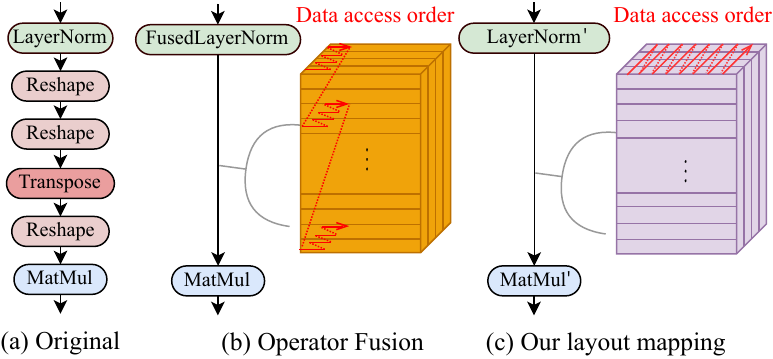}
  \caption{{\bf Data access patterns comparison before and after layout and access optimizations.} \revisioncr{In left (a), two consecutive {\tt Shape} operations reshape the output from {\tt LayerNorm} across different dimensions.}}
  \label{fig:example_layout_access_dim}
\end{figure}

$L^1$ and $L^2$ do not have any reduction dimension because they are consumed by an element-wise addition operator ({\tt Add}), so theoretically we are allowed to map either D1/D3 or D2 to that 0.5 dimension of 2.5D memory. 
However, because $L^1$ and $L^2$ are used together with $L^0$ in the {\tt FusedConv} and their element-wise addition operations have been fused with the {\tt Conv}, \projectname maps them to 2.5D memory in a similar manner to $L^0$ (i.e., mapping D1 and D3 to the 0.5D of 2.5D memory, respectively), avoiding extra index computations.

\noindent\underline{\em Optimized memory access based on tensor layout.} \xspace\xspace Figure~\ref{fig:example_layout_access_dim}
shows the high-level idea of our optimized data access on 2.5D memory by comparing the data access orders before and after this optimization. 
Figure ~\ref{fig:example_layout_access_dim} 
(a) shows the original computational graph, Figure ~\ref{fig:example_layout_access_dim} 
(b) shows the computational graph and data access order after the fusion and operator elimination offered by \projectname, and Figure ~\ref{fig:example_layout_access_dim} 
(c) shows the computational graph and data access order after our optimized tensor layout mapping and memory access optimization. 
Before the optimized tensor layout mapping, although the {\tt Reshape} and {\tt Transpose} operations are fused (and eliminated), the memory access pattern is complex and fragmented, resulting in poor data locality (and similarly low SIMD efficiency). 
The optimized tensor layout mapping offers us an opportunity to access this tensor along its reduction dimension with a stride of 1,  thus improving the data locality on 2.5D memory and cache, and parallel and SIMD efficiency on mobile GPUs.

\noindent\underline{\em Other optimizations.}
In addition to the previously mentioned optimizations, our framework incorporates an auto-tuning mechanism that utilizes Genetic Algorithms~\cite{niu2021dnnfusion} 
for generating GPU execution configurations. These configurations include block dimensions, unrolling factors, and tiling shapes.

\begingroup
\setlength{\tabcolsep}{1.7pt}
\begin{table*}[t!]
     
  \centering
  \caption{\revisioncr{{\bf Model characterization and comparing the number of operators among frameworks}. Hybrid means the combination of Transformer and ConvNet. ``--'' means the model is not supported by the framework yet.}}
  \label{tab:eva_model_info}
  \small
  \begin{tabular}{lrrr|rrr|rrrrrr}
    \toprule
    \multirow{2}{*}{Model} & \multirow{2}{*}{Model Type} & \multirow{2}{*}{Input Type} & \multirow{2}{*}{Attention} & \multirow{2}{*}{\#Operators$^\dagger$} & \multirow{2}{*}{\#Params (M)} & \multirow{2}{*}{\#MACs (G)} & \multicolumn{6}{c}{\#Operators with optimizations}\\\cline{8-13}
    ~                                       & ~           & ~ & ~      & ~    & ~     & ~   & MNN   & NCNN & TFLite & TVM & DNNF & Ours \\\hline
    AutoFormer~\cite{AutoFormer,S3}         & Transformer & Image & Local  & 546  & 31.2  & 4.7  & 449   & --  & --   & 302  & 197 & {\bf 148} \\ 
    BiFormer~\cite{zhu2023biformer}         & Hybrid      & Image & Local  &2,042 & 25.5  & 4.5  & 1,189 & --  & --   & 1,029& 602 & {\bf 474} \\ 
    CrossFormer~\cite{wang2021crossformer}  & Transformer & Image & Local  & 505  & 31    & 5    & 453   & --  & --   & 308  & 196 & {\bf 155} \\ 
    CSwin~\cite{dong2021cswin}              & Hybrid      & Image & Local  &3,863 & 34.7  & 6.9  & 1,753 & --  & --   & 1,480& 933 & {\bf 604} \\ 
    EfficientVit~\cite{cai2022efficientvit} & Hybrid      & Image & Local  & 536  & 51    & 5.2  & 489   & --  & --   & 133  & 113 & {\bf 101} \\ 
    FlattenFormer~\cite{han2023flatten}     & Hybrid      & Image & Local  &2,016 & 37.3  & 7.2  & 1,558 & --  & --   & 918  & 665 & {\bf 403} \\ 
    SMTFormer~\cite{lin2023scaleaware}      & Hybrid      & Image & Local  &1,406 & 22.5  & 4.9  & 1,905 & --  & --   & 844  & 469 & {\bf 332} \\ 
    Swin~\cite{liu2021Swin}                 & Transformer & Image & Local  & 765  & 28.9  & 4.6  & 596   & --  & --   & 374  & 207 & {\bf 158} \\ 
    ViT~\cite{dosovitskiy2020image}         & Transformer & Image & Global & 444  & 102.8 & 21   & 379   & --  & --   & 289  & 168 & {\bf 112} \\
    \midrule
    Conformer~\cite{gulati2020conformer}    & Hybrid      & Audio & Global & 665  & 10    & 12   & 558   & --  & --   & 356  & 219 & {\bf 163} \\
    SD-TextEncoder~\cite{rombach2022high}   & Transformer & Text  & Global & 674  & 123   & 6.7  & 601   & --  & --   & 297   & 101 & {\bf  84} \\
    SD-UNet~\cite{rombach2022high}          & Hybrid      & Image & Global &1,962 & 860   & 91   & 1355  & --  & --   & 889  & 436 & {\bf 322} \\
    SD-VAEDecoder~\cite{rombach2022high}    & Hybrid      & Image & Global & 287  & 50    & 312  & 206   & --  & --   & 156  & 103 & {\bf  95} \\
    Pythia~\cite{biderman2023pythia}        & Transformer & Text  & Decoder&1,853 &1,121  & 119  & 809   & --  & --   & 681  & 525 & {\bf 355} \\
    \midrule
    ConvNext~\cite{liu2022convnet}          & ConvNet     & Image & N/A    & 292  & 28.6  & 4.5  & 321   & --  & --   & 185  & 96  & {\bf 81} \\
    RegNet~\cite{radosavovic2020designing}  & ConvNet     & Image & N/A    & 282  & 19.4  & 3.2  & 197   & 282 & 197  & 155  & 122 & {\bf 122} \\
    ResNext~\cite{xie2017aggregated}        & ConvNet     & Image & N/A    & 122  & 25    & 4.3  & 86    & 122 & 73   & 58   & 55  & {\bf 55} \\
    Yolo-V8~\cite{yolov8_ultralytics}       & ConvNet     & Image & N/A    & 233  & 3.2   & 4.4  & 176   & 233 & --   & 88   & 75  & {\bf 68} \\
    \bottomrule
    \multicolumn{13}{l}{\footnotesize \#Operators$^\dagger$ denotes the number of operators in the unoptimized computational graph.} \\
  \end{tabular}
\end{table*}
\endgroup

\section{Evaluation}\label{sec:eval} 

This section evaluates the \projectname by comparing it to five state-of-the-art frameworks across different models. The objectives of this evaluation are as follows: 
1) To exhibit the notable improvements \projectname offers against existing, cutting-edge DNN frameworks on a 
\revisioncr{mobile} GPU, 
2) To explore how various optimizations contribute to these performance enhancements, 
3) To demonstrate the portability of proposed optimizations in \projectname by evaluating the execution times 
on two other mobile platforms,  and 
4) To illustrate that our proposed optimizations enable performance improvement on a desktop-level GPU, 
which is less resource-constrained and has a traditional (one-dimensional) memory. 
Particularly, \projectname is compared against MNN~\cite{Ali-MNN}, NCNN~\cite{Ni_ncnn_2017}, TFLite~\cite{TensorFlow-Lite}, TVM~\cite{chen2018tvm}, and DNNFusion~\cite{niu2021dnnfusion}(refers as DNNF), on the mobile GPU.

\subsection{Evaluation Setup}

\noindent{\bf DNN Workloads:}
Our evaluation is conducted on 18 state-of-the-art DNN models with three different structures, including Transformer, ConvNet, and Hybrid (i.e., ones having both Transformer and ConvNet structures), as well as Stable Diffusion and LLMs. 
Table~\ref{tab:eva_model_info} characterizes them with a comparison of their type, attention mechanism, the number of parameters, the number of operators prior to optimizations, as well as the number of multiply-accumulate operations (MACs). 
\revisioncr{
We have 1) six Transformer models (AutoFormer~\cite{AutoFormer,S3}, CrossFormer~\cite{wang2021crossformer}, Swin~\cite{liu2021Swin}, ViT~\cite{dosovitskiy2020image}, Stable Diffusion - TextEncoder~\cite{rombach2022high}, Pythia~\cite{biderman2023pythia}), 2) four Convolution models (ConvNext~\cite{liu2022convnet}, RegNet~\cite{radosavovic2020designing}, ResNext~\cite{xie2017aggregated}, Yolo-V8~\cite{yolov8_ultralytics}), and 3) eight Hybrid models with both Transformer and Convolution structures (BiFormer~\cite{zhu2023biformer}, CSwin~\cite{dong2021cswin}, EfficientVit~\cite{cai2022efficientvit}, FlattenFormer~\cite{han2023flatten}, SMTFormer~\cite{lin2023scaleaware}), Conformer~\cite{gulati2020conformer}, StableDiffusion - UNet and VAEDecoder~\cite{rombach2022high}.
}

The Transformer structure can serve as a backbone for different CV and NLP tasks. Due to the space limitations, we report the results on the object detection task (for Yolo-V8) 
 and image classification task (for all other models) in this evaluation.
Since the training dataset has a negligible impact on the final inference latency,
this section reports results from one training dataset for each model.
Yolo-V8 is trained on MS COCO dataset~\cite{cocodataset};
Conformer, Stable Diffusion Models (SD-TextEncoder, SD-UNet, SD-VAEDecoder), and Pythia are from the pretrained checkpoints;
other models are trained on ImageNet dataset~\cite{deng2009imagenet}.
Since the model accuracy is the same across all frameworks, our evaluation focuses only on execution latency.

\noindent{\bf Evaluation environment.}
\projectname is built upon \revisioncr{our previous work} (DNNFusion~\cite{niu2021dnnfusion}) that supports extensive operator fusion. 
We conduct our major experiments on a Oneplus cell phone using the high-end Qualcomm Snapdragon 8 Gen 2 platform~\cite{snapdragon8gen2}, which includes a Qualcomm Kryo octa-core CPU 
and a Qualcomm Adreno 740 GPU with 16 GB of unified memory (shared by both CPU and GPU).
In addition, to demonstrate our portability, we test \projectname on an earlier generation of the Qualcomm platform - Snapdragon 835~\cite{snapdragon835}, which has more limited resources, and on a MediaTek platform with Dimensity 700 SOCs.
The Snapdragon 835 consists of an ARM Octa-core CPU, an Adreno 540 GPU, and 6 GB of unified memory. The MediaTek Dimensity 700 is equipped with an ARM Octa-core CPU, a Mali-G57 GPU, and 4 GB of unified memory.
Furthermore, we evaluate our optimizations on an NVIDIA GPU to illustrate our generality. For 
this comparison, we implement our optimizations (excluding the mobile device-specific optimization for the 2.5D memory layout) in TorchInductor and compare it against the base version of TorchInductor. 
For all evaluated models and frameworks on mobile devices, GPU execution uses 16-bit floating-point representation. On desktop-level GPU, we use 32-bit floating-point representation for evaluation purposes. 
\revisioncr{Noting that, we use 16-bit floating-point for mobile GPU because it is a common data type that all the frameworks support. Although other data types may have different accuracies, our optimizations are based on operator semantics thus not limited to specific data types.}
\revisioncr{The batch size is set to 1 for all models unless otherwise specified.}
With the results we reported in this section, we have utilized the auto-tuning capabilities available in MNN, TVM, and TorchInductor to achieve the best possible performance. 
Each experiment is executed 50 times and only the average numbers are reported – as the variance was negligible, it is omitted for readability.

\begingroup
\setlength{\tabcolsep}{3.7pt}
\begin{table*}[t!]
  \centering
  \caption{{\bf End-to-end latency comparison across different frameworks -- execution on GPU of Snapdragon 8 Gen 2. `--' means that the model is not supported. `SD' represents for StableDiffusion, which includes three pipelines.}}
  \label{tab:eva_model_eval}
  \small
  \begin{tabular}{lr|rrrrrr|rrrrrr|c}
    \toprule
    \multirow{2}{*}{Model} & \#MACs & \multicolumn{6}{c|}{Latency (ms)} & \multicolumn{6}{c|}{Speed (GMACS)} & {\bf Speedup} \\ 
    ~              & (G)  & MNN   & NCNN & TFLite & TVM   & DNNF & {\bf Ours} & MNN & NCNN & TFLite & TVM   & DNNF & {\bf Ours} &  {\bf over DNNF} \\ \hline
    AutoFormer     & 4.7  & 335   & --   & --     & 184   & 106  & 40.2       & 14  & --   & --     & 25    & 44   & 117 & \revisioncr{2.6$\times$} \\
    BiFormer       & 4.5  & 1,736 & --   & --     & 208   & 186  & 56.1       & 3   & --   & --     & 22    & 24   & 81  & \revisioncr{3.3$\times$} \\
    CrossFormer    & 5    & 336   & --   & --     & 156   & 121  & 38.2       & 15  & --   & --     & 32    & 41   & 130 & \revisioncr{3.2$\times$} \\
    CSwin          & 6.9  & 703   & --   & --     & 261   & 225  & 57.6       & 10  & --   & --     & 26    & 31   & 120 & \revisioncr{3.9$\times$} \\
    EfficientVit   & 5.2  & 208   & --   & --     & 243   & 112  & 22.5       & 25  & --   & --     & 21    & 47   & 232 & \revisioncr{5.0$\times$} \\
    FlattenFormer  & 7.2  & 492   & --   & --     & 256   & 210  & 60.1       & 15  & --   & --     & 28    & 34   & 119 & \revisioncr{3.5$\times$} \\
    SMTFormer      & 4.9  & 510   & --   & --     & 214   & 143  & 40         & 10  & --   & --     & 23    & 35   & 124 & \revisioncr{3.6$\times$} \\
    Swin           & 4.6  & 372   & --   & --     & 158   & 135  & 30.6       & 15  & --   & --     & 29    & 34   & 149 & \revisioncr{4.4$\times$} \\
    ViT            & 21   & 533   & --   & --     & 1,050 & 277  & 103        & 39  & --   & --     & 20    & 76   & 204 & \revisioncr{2.7$\times$} \\
    \midrule
    Conformer      & 12   & 1,736 & --   & --     & 863   & 284  & 106        & 7   & --   & --     & 14    & 42   & 113 & \revisioncr{2.7$\times$}  \\
    SD-TextEncoder & 6.7  & 153   & --   & --     & 216   & 73   & 38         & 44  & --   & --     & 31    & 92   & 176 & \revisioncr{1.9$\times$}  \\
    SD-UNet        & 90   & 2,172 & --   & --     & 3,969 &1,108 & 412        & 42  & --   & --     & 23    & 82   & 219 & \revisioncr{2.7$\times$}  \\
    SD-VAEDecoder  & 312  & 2,730 & --   & --     & 5,663 &1,596 & 866        & 114 & --   & --     & 55    & 195  & 360 & \revisioncr{1.8$\times$}  \\
    Pythia         & 119  & 3,034 & --   & --     & 6,602 &1,489 & 663        & 40  & --   & --     & 18    & 80   & 180 & \revisioncr{2.3$\times$}  \\
    \midrule
    ConvNext       & 4.5  & 271   & --   & --     & 5,543 & 109  & 33.4       & 17  & --   & --     & 0.81  & 41   & 135 & \revisioncr{3.3$\times$} \\
    RegNet         & 3.2  & 61    & 33   & 36.4   & 71    & 31   & 24.7       & 52  & 106  & 88     & 45    & 103  & 129 & \revisioncr{1.3$\times$} \\
    ResNext        & 4.3  & 158   & 38   & 66     & 106   & 33   & 15.7       & 27  & 111  & 64     & 40    & 129  & 271 & \revisioncr{2.1$\times$} \\
    Yolo-V8        & 4.4  & 32    & 28   & --     & 141   & 26   & 22         & 138 & 169  & --     & 31    & 169  & 200 & \revisioncr{1.2$\times$} \\\midrule
    \multicolumn{2}{c|}{Geo-mean (speedup)}      & \revisioncr{7.9$\times$} & \revisioncr{1.6$\times$} & \revisioncr{2.5$\times$} & \revisioncr{6.9$\times$} & \revisioncr{2.8$\times$} & \revisioncr{1.0$\times$} & \revisioncr{7.9$\times$} & \revisioncr{1.6$\times$} & \revisioncr{2.5$\times$} & \revisioncr{6.9$\times$} & \revisioncr{2.8$\times$} & \revisioncr{1.0$\times$} & - \\
    \bottomrule
    \multicolumn{15}{l}{\footnotesize The results collected from MNN/TVM/DNNFusion with auto-tuning.} \\
  \end{tabular}
\end{table*}
\endgroup

\subsection{Overall Performance Comparison}

\noindent\textbf{Fusion rate comparison.}
Table~\ref{tab:eva_model_info} shows the model information with and without optimizations, including the number of operators in the unoptimized models as well as the  number of operators in optimized versions produced by 
each framework (``--'' indicates that the model is not supported by the framework). 
NCNN and TFLite do not support Transformer models on mobile GPU as they either lack support for key operators and/or do not reduce the memory requirements sufficiently. 
\projectname achieves higher fusion rates compared to MNN and TVM for Transformer and Hybrid models, with ratios ranging from \revisioncr{ $2.2\times$ to $7.2\times$, and $1.3\times$ to $3.5\times$ }, respectively.
For ConvNet models, \projectname achieves fewer operators as compared to 
MNN, NCNN, TFLite, and TVM by a factor of \revisioncr{ $1.6\times$ to $4.0\times$}, $2.2\times$ to $3.4\times$, $1.3\times$ to $1.6\times$, and $1.1\times$ to $2.3\times$, respectively.
\projectname offers greater benefits for Transformer and Hybrid models (e.g., BIformer, CSwin, SMTFormer, ViT) compared to ConvNet models (ResNext, RegNet, Yolo-V8). 
This is because Transformer/Hybrid models require more frequent data reshaping and transposing.  
The significant improvement in fusion rate stems from two separate components in \projectname: 
the optimization based on DNNFusion which is achieving higher fusion rates compared to other frameworks, 
and the novel (reduction dimension-based elimination optimizations) introduced in this paper. 
Separating the two, we note that  \projectname achieves a fusion rate of up to \revisioncr{$1.7\times$} compared to DNNFusion.
For three of the  ConvNets (RegNet, ResNext, Yolo-V8), \projectname demonstrates a similar fusion rate as DNNFusion  
as there are fewer Reshape and Transpose operations present. 
However, for the more complex ConvNet model ConvNext, \projectname achieves an additional fusion opportunity of \revisioncr{$1.2\times$} compared to DNNFusion.
Furthermore, for Transformer and Hybrid models,\projectname enables \revisioncr{$1.1\times$ to $1.7\times$} more fusion opportunities through our elimination techniques when compared to DNNFusion.

\noindent\textbf{End-to-end execution time comparison.}
Table~\ref{tab:eva_model_eval} presents a comparison of overall execution latency and speed (measured in GMACS) among six frameworks on our target mobile GPU. In the case of Transformer and Hybrid models, 
\projectname achieves $3.2\times$ to $30.9\times$, and $3.7\times$ to $10.8\times$ speedups compared with MNN and TVM,  respectively (NCNN and TFLite do not support Transformer and Hybrid models on mobile GPU). 
As for ConvNet models,
\projectname achieves $1.5\times$ to $10.1\times$, $1.3\times$ to $2.4\times$, $1.5\times$ to $4.2\times$, and $2.9\times$ to $166\times$ compared with MNN, NCNN, TFLite, and TVM, respectively.
The exceptionally large speedup on ConvNext when compared to TVM is due to TVM lacking an efficient layout design for a reduction operator  {\tt GroupConvolution}. 
\revisioncr{For BIFormer, \projectname significantly outperforms MNN because the token selection mechanism in BIFormer leads to more data transformation operations, resulting in more benefits from eliminating layout transformations.}
Compared to DNNFusion, \projectname achieves $1.8\times$ to $5.0\times$ speed improvement for Transformer and Hybrid models, while achieving $1.2\times$ to $3.3\times$ improvement for ConvNets.
The gains on ConvNets compared with DNNFusion are because our layout selection enables more hardware-friendly data access and computation pattern across multiple operators.
It is worth noting that \projectname delivers similar speed (around 120 GMACS) for models with similar structures regardless of the number of operators, i.e., for AutoFormer, BIFormer, CrossFormer, CSwin, FlattenFormer, SMTFormer, and Swin. 
On the other hand, for EfficientViT, ViT, ResNext, and Yolo-V8 models, the speed is higher (over 200 GMACS). This is because these models either have more intensive computation in each operator (e.g., ViT and EfficientViT), which means more data reuse in computation or have a more regular computation pattern (RegNet and ResNext) with pure Convolution structure.

\begin{figure}[t]
  \centering
  \subfloat[Memory access count]{\includegraphics[width=0.49\columnwidth]{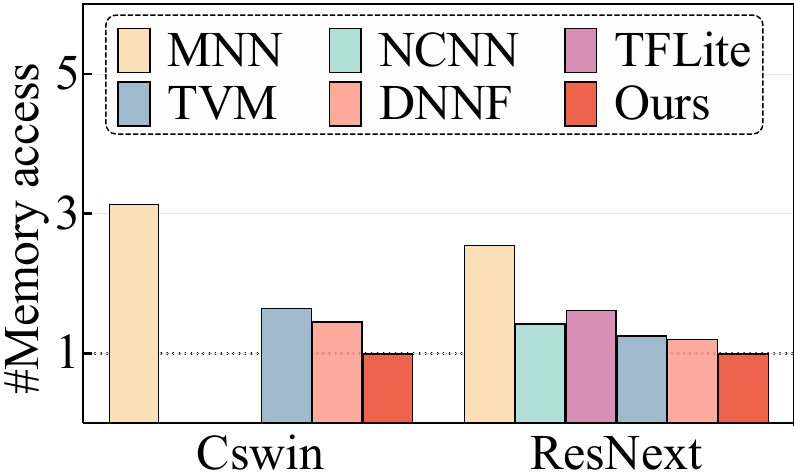}}
  \subfloat[Cache miss count]{\includegraphics[width=0.49\columnwidth]{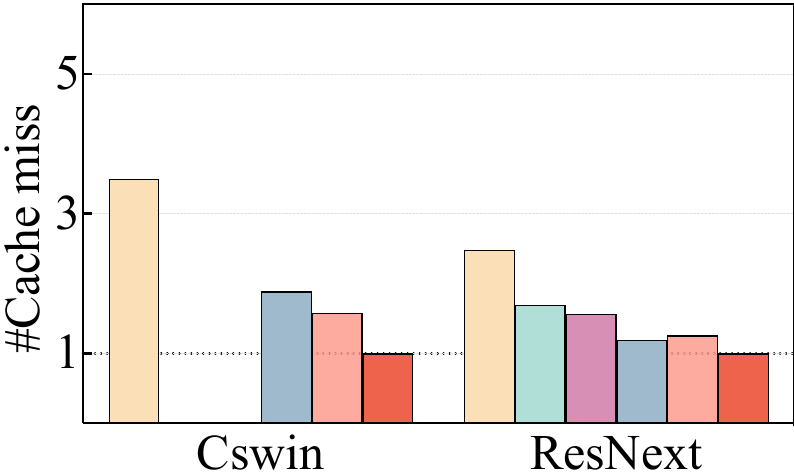}}
  \caption{{\bf Memory access count and cache miss count comparison compared with other frameworks.} \revisioncr{The memory access data is collected from the hardware counter on the mobile GPU, showing the total number of data accesses happening in the global memory of the mobile GPU.} All results are normalized by ours.}
  \label{fig:eva_overall_memory_access}
\Description[]{}
\end{figure}

\noindent\textbf{Memory performance (overall).}
Figure~\ref{fig:eva_overall_memory_access}(a) compares the number of memory accesses (left) and the number of cache misses (right). 
The results are normalized by {\tt Ours} (\projectname)  for readability.
Due to space limitations, we only present results for one local attention Hybrid model 
(Cswin) and one ConvNet model (ResNext). 
Cswin execution using our framework is compared against MNN, TVM, and DNNFusion, whereas ResNext execution is compared against 
these as well as NCNN and TFLite. 
As shown in Figure ~\ref{fig:eva_overall_memory_access}, \projectname utilizes 1.8$\times$ fewer memory accesses on average and achieves an average of 2.0$\times$ fewer cache misses compared to other frameworks.
In the following section, we further study the impact of our key optimizations on memory accesses.

\begin{figure}[t]
  \centering
  \includegraphics[width=0.92\columnwidth]{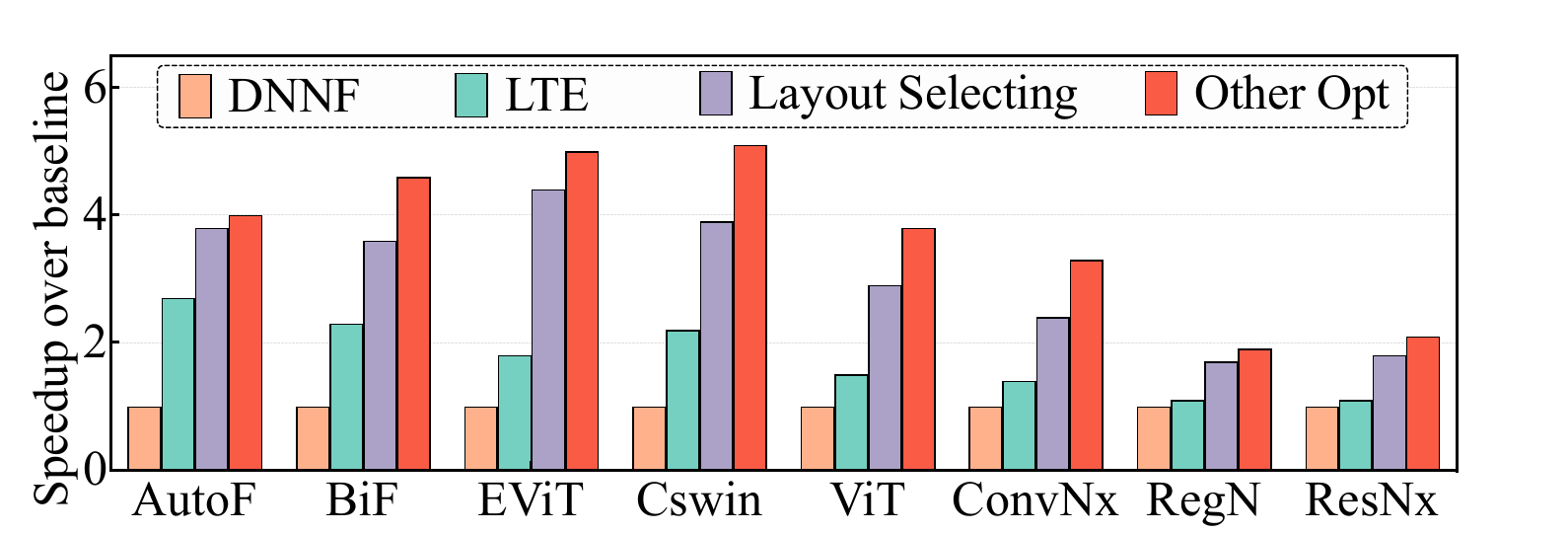}
  \caption{{\bf Performance breakdown analysis}: speedup over baseline (DNNFusion).}
  \label{fig:eva_speedup_breakdown}
  \Description[]{}
\end{figure}

\subsection{Optimization Breakdown and Analysis}
This section studies the impact of the key optimizations in \projectname on execution latency. We evaluate  
the gains from each optimization incrementally over our baseline version DNNF (shorts for DNNFusion).
Figure ~\ref{fig:eva_speedup_breakdown} illustrates the impact of our proposed optimizations on latency with eight models. 
These include five Transformer and Hybrid models (AutoFormer (AutoF), BiFormer (BiF), EfficientViT (EViT), Cswin, ViT) and three ConvNet models (ConvNext(ConvNx), RegNet (RegN), ResNext (ResNx)) on the mobile GPU.
LTE shorts for Layout Transformation Elimination.
Due to space limitations, we omit experiments on other models that show a similar trend.
For Transformer and Hybrid, compared to DNNF, Layout Transformation Elimination (LTE) achieves a speedup of 1.5$\times$ to 2.7$\times$.
Layout Selecting brings an additional speedup of 1.4$\times$ to 1.9$\times$,
and other optimizations (texture memory-related and tuning) bring an additional speedup of 1.2$\times$ to 1.4$\times$.
For ConvNet, these numbers are 1.1$\times$ to 1.4$\times$, 1.5$\times$ to 1.7$\times$, and 1.1$\times$ to 1.4$\times$, respectively.
\revisioncr{Considering Index Comprehension, for Transformer and Hybrid models, it contributes to the speedup results of Layout Transformation Elimination by 1.1$\times$ to 1.3$\times$, while for ConvNets, it contributes between 1.1$\times$ and 1.2$\times$.}

\begin{figure}[t]
  \centering
  \subfloat[Memory access counts]{\includegraphics[width=0.47\columnwidth]{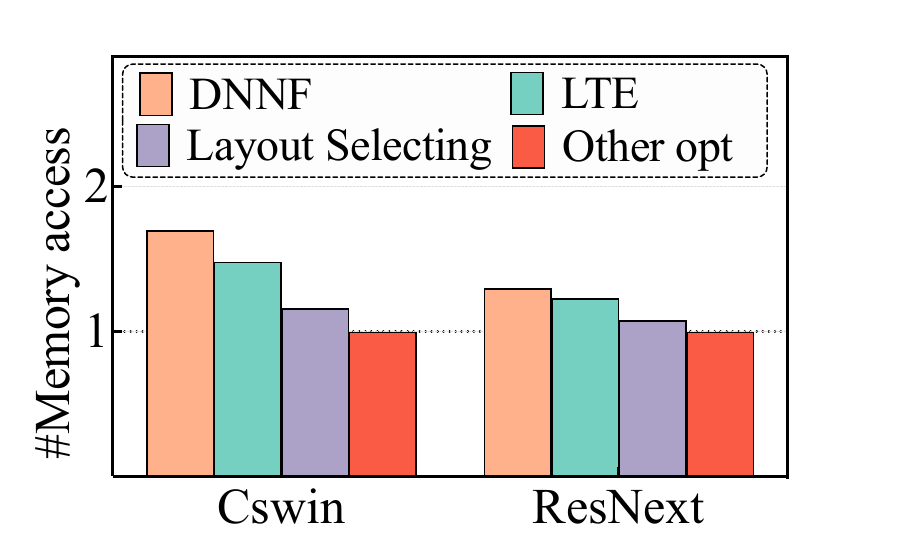}}
  \subfloat[Cache miss counts]{\includegraphics[width=0.47\columnwidth]{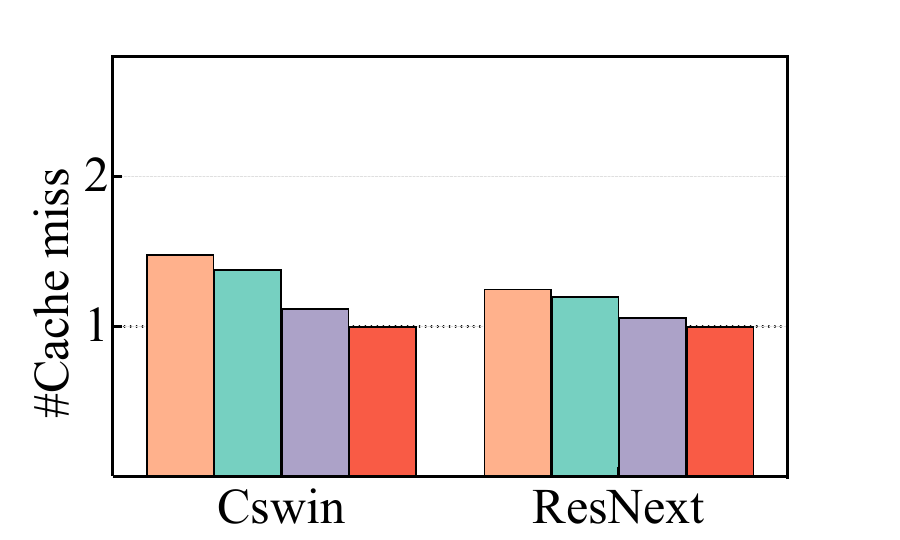}}
  \caption{{\bf Optimization breakdown: memory access count and cache miss count comparison}.}
  \label{fig:eva_memory_access}
\Description[]{}
\end{figure}

\noindent\textbf{Memory and cache breakdown.}
To further investigate the impact of our key optimizations on the underlying hardware, Figure ~\ref{fig:eva_memory_access} illustrates the breakdown of memory and cache usage in \projectname.
Figure ~\ref{fig:eva_memory_access} demonstrates the reduction in memory access counts (left) and cache miss counts (right) for Cswin and ResNext, respectively.
Layout Transformation Elimination has a greater effect on reducing memory access count than cache miss count because it eliminates the need for data reorganization and enables more fusion opportunities. 
On the other hand, Layout Selection has a greater impact on cache miss count than memory access count as it designs better layouts for each operator thus enabling better data access patterns.

\begin{figure}[t]
  \centering
  \subfloat[Swin]{\includegraphics[width=0.92\columnwidth]{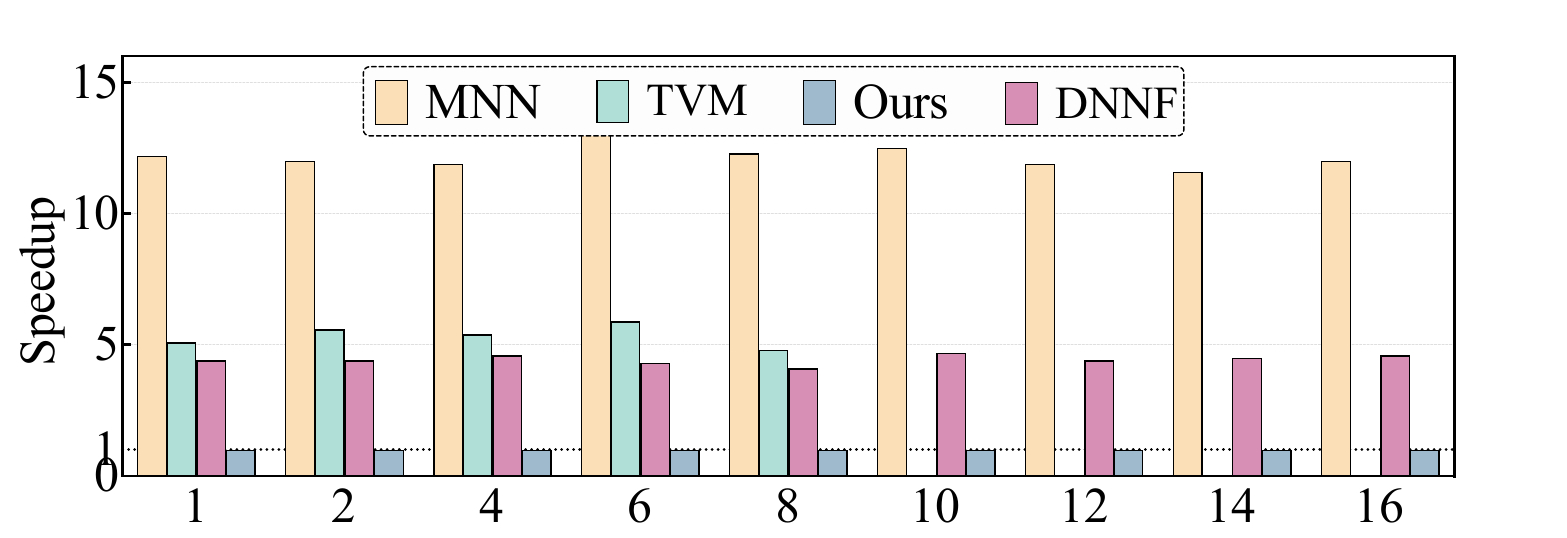}}
  \caption{\revisioncr{{\bf Performance comparison with different batch sizes for Swin.} We report the speedup for better readability. }}
  \label{fig:eva_batch_mobile}
\Description[]{}
\end{figure}

\begin{figure}[t]
  \centering
  \subfloat[On Mali-G57 (Mediatek Dimensity 700)]{\includegraphics[width=0.92\columnwidth]{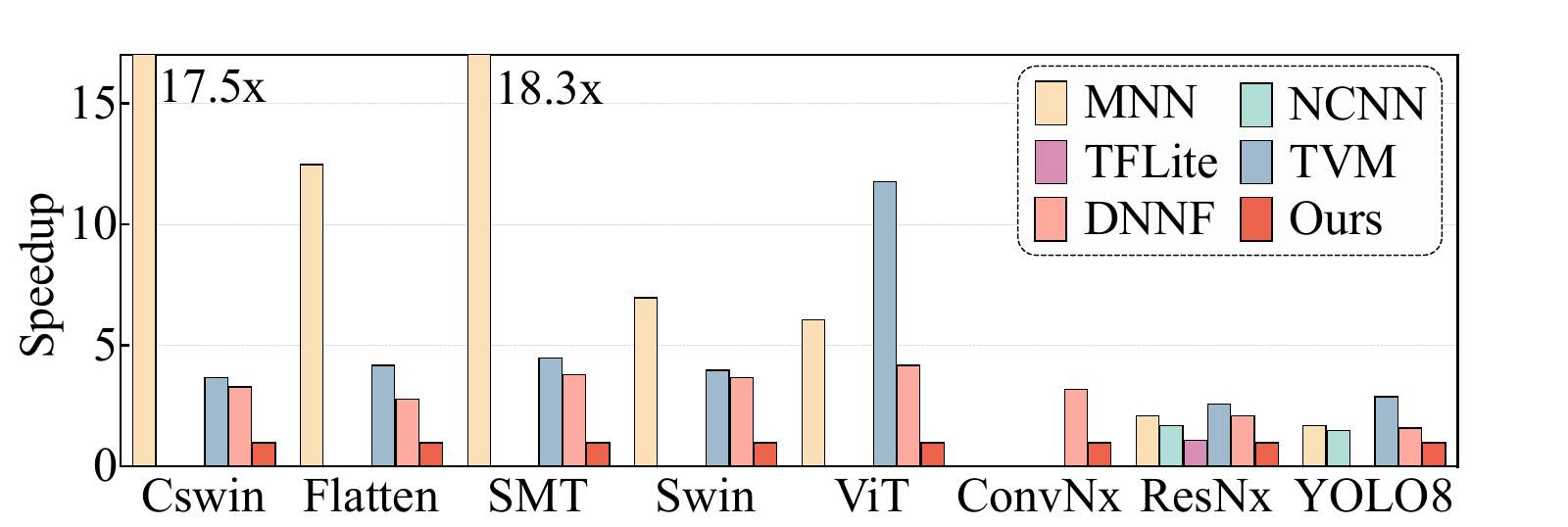}}
  \quad 
  \subfloat[On Adreno 540 (Qualcomm Snapdragon 835)]{\includegraphics[width=0.92\columnwidth]{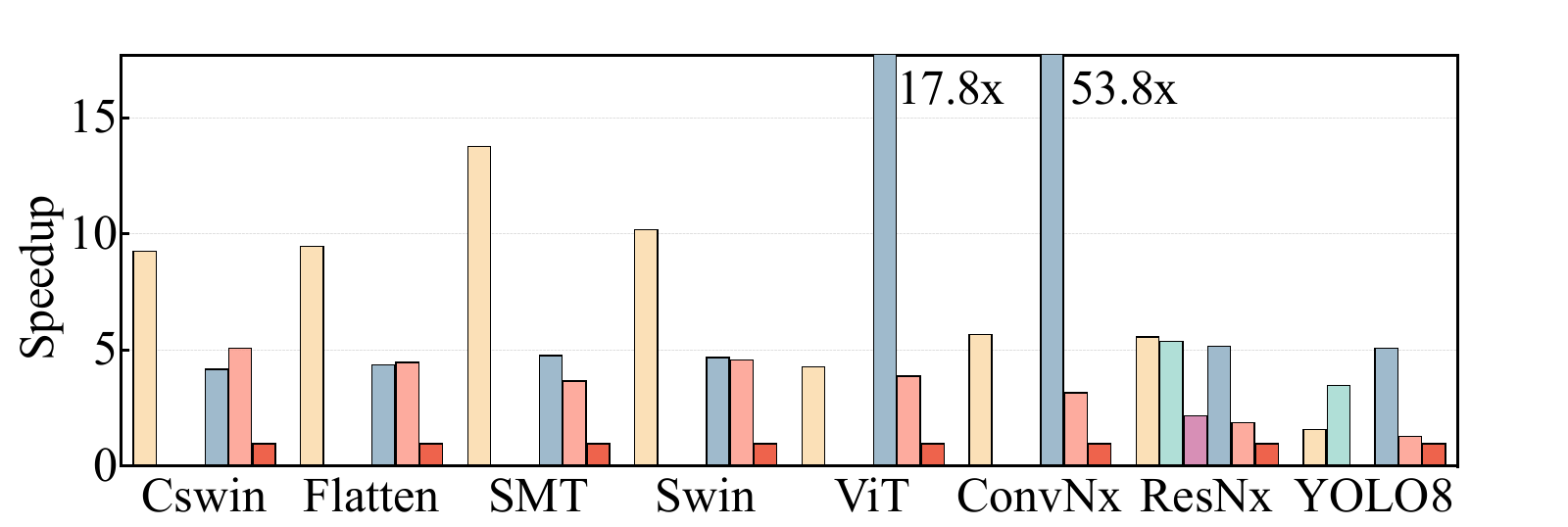}}
  \caption{{\bf Portability evaluation on Mediatek Dimensity 700 and Qualcomm Snapdragon 835}. We report the speedup for better readability. Empty bar on other frameworks means the model is not supported due to lack of operator supports or insufficient memory on device. }
  \label{fig:eva_portability}
\Description[]{}
\end{figure}

\revisioncr{\subsection{Performance Impact of Batch Size}
Figure ~\ref{fig:eva_batch_mobile} illustrates the performance improvements achieved by \projectname on mobile GPU, compared to MNN, TVM, and DNNF, across different batch sizes. 
We report the results for a representative Transformer model, Swin with batch sizes ranging from 1 to 16. Other models show a similar trend.
An empty bar for other frameworks indicates that the corresponding batch size is not supported due to insufficient device memory. 
For different batch sizes, \projectname shows speed improvements of 11.6$\times$ to 13.2$\times$, 4.8$\times$ to 5.9$\times$, and 4.1$\times$ to 4.7$\times$ compared to MNN, TVM, and DNNFusion, demonstrating its scalability.}

\subsection{Portability Evaluation} 
\noindent\textbf{Older Mobile Platforms:} 
Figure ~\ref{fig:eva_portability} shows the speedup achieved on  Mediatek Dimensity 700 and Qualcomm Snapdragon 835.  
MNN and TVM do not support ConvNext on Mali-G57 due to insufficient memory (4GB unified memory).
 \projectname achieves similar speedup on these platforms with limited resources, and is actually 
 less sensitive to having fewer resources  because our Layout Transformation Elimination and Layout Selection reduces the overall number of operators, resulting in lower memory and cache pressure.

\noindent\textbf{Desktop-level GPU:}
We have implemented our Layout Transformation Elimination and Layout Selecting on TorchInductor and conducted a performance comparison.
TorchInductor automatically selects the optimal implementation from either TensorRT or a Pytorch built-in or compiler-generated Triton kernel. 
We compared the performance of two Transformer models, Swin and AutoFormer, in this environment. 
As shown in Table~\ref{tab:eva_desktop_gpu}, \projectname, achieves $1.23\times$ and $1.11\times$ better execution efficiency than TorchInductor with data type FP32.
As these results are from a quick and separate implementation, we expect that additional benefits will be possible in future work.

\begingroup
\setlength{\tabcolsep}{2.6pt}
\begin{table}[t!]
\centering
\caption{{\bf Comparison of end-to-end execution latency on NVIDIA GPU between PytorchInductor and \projectname for Swin and SMTFormer.} Results are collected with a batch size of 1.}
\footnotesize{
\begin{tabular}{ccc|cc}
    \toprule
    Model & System    & Device  & TorchInductor (ms) & Ours (ms) \\
    \hline
    Swin       & Ubuntu 20.04 & Tesla V100 & 7.5 & 6.1  \\ 
    AutoFormer  & Ubuntu 20.04 & Tesla V100 & 5.1 & 4.6 \\ 
    \bottomrule
\end{tabular}
}
\label{tab:eva_desktop_gpu}
\end{table}
\endgroup 

\subsection{Discussion}

\noindent\textbf{Limitations.}
\noindent\revisioncr{While our optimizations significantly improve the performance of transformer models with many layout transformation operations, it is worth pointing out that \projectname yields only modest performance improvements for traditional convolutional neural networks such as Yolo-V8 and RegNet, or models in general with few layout transformation operations compared to DNNF.
In comparison to DNNF (the best baseline shown in Table~\ref*{tab:eva_model_eval}) in CNN-only models, particularly RegNet and Yolo-V8, \projectname achieves a speedup of 1.3$\times$ and 1.2$\times$, respectively. These two models primarily consist of computationally intensive operators (Conv2D), and the advanced operator fusion proposed in DNNF works effectively.
Conversely, for CNNs with more layout transformation operations (ConvNext) or those sensitive to the data layouts like ResNext, \projectname delivers a speedup of 3.3$\times$ and 2.1$\times$, demonstrating the effectiveness of our optimization in eliminating layout transformations and selection for these models.
}

\begin{figure}[t]
  \centering
  \includegraphics[width=0.92\columnwidth]{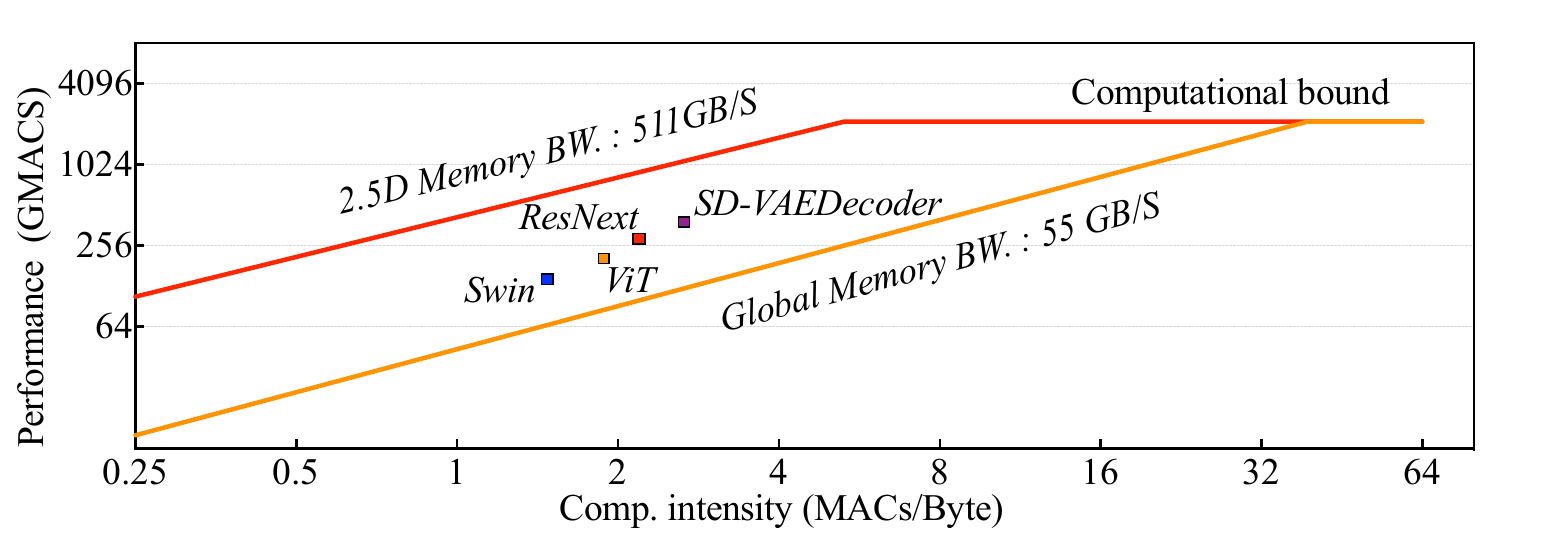}
  \caption{Roofline analysis for peak and actual performance (GMACs/s) on mobile GPU for ViT. 
  }
  \label{fig:eva_roofline_vit}
  \Description[]{}
\end{figure}

\noindent\textbf{Roofline analysis for the achieved performance.}
\noindent\revisioncr{
To better evaluate \projectname's optimization effect,
Figure~\ref{fig:eva_roofline_vit} shows the results from roofline analysis~\cite{williams2009roofline} on four models (Swin, ViT, ResNext, and SD-VAEDecoder). Here, the yellow line and red line show the theoretical peak performance assuming that all data is accessed from global memory and 2.5D texture memory, respectively. The global memory and texture memory bandwidths are 55 GB/s and 511 GB/s, respectively, and the mobile GPU's peak computation performance is 2.0 TMACs/s~\cite{liang2022romou}.
These models consist of various operators with different computation intensity levels. 
Certain operators have a low computation intensity (e.g., activation functions and elementwise arithmetic operations),
while others exhibit a high computation intensity (e.g., {\tt Conv} and {\tt MatMul}).
For ease of comparison, the x-axis in this analysis denotes the computation intensity (MACs/byte)  
averaged across the entire computational graph.} 

\noindent\revisioncr{
Figure~\ref{fig:eva_roofline_vit} shows the analysis for Swin, ViT, ResNext, and SD-VAEDecoder, which all have 
different levels of computational intensity, achieve 149, 204, 271, and 360 GMACs/s, respectively. 
Specifically, Swin performs the worst while SD-VAEDecoder performs the best because Swin contains more low computation intensity operators whereas  SD-VAEDecoder contains more high computation intensity operators.  
Under an unrealistic assumption that all data accesses are from  2.5D texture memory,  \projectname achieves 24\%, 27\%, 31\%, and 35\% of the theoretical peak performance. A higher computation intensity of a model implies more {\tt Conv} and/or {\tt MatMul} operators in the model. 
These operators are more likely to read data from 2.5D texture memory, thus achieving a higher percentage of the theoretical peak performance.
}

\noindent\textbf{Impact of redundant data copies with different layouts.}
\revisioncr{
The \projectname stores redundant copies of data, specifically intermediate results, 
with different layouts when an operator has multiple consumers requiring varied input data layouts. 
During runtime, these redundant layouts do not significantly increase memory consumption for the models we evaluated. 
This is primarily due to two reasons. 
Firstly, maintaining multiple copies of data in different layouts is not commonly seen during execution; for instance, Swin and ViT have a maximum active redundant copy size of 3.0MB and 2.3MB, respectively. 
Additionally, similar to TVM, our implementation allocates intermediate results from a memory pool allowing efficient reuse of memory resources by releasing data copies back into the pool when they are no longer needed by any consumers.
In addition, compared to DNNFusion, eliminating kernels reduces further memory usage. Take SwinTransformer and ViT as examples: \projectname decreases the number of operators by 24\% and 33\% compared to DNNF, as shows in Table~\ref{tab:eva_model_info}, resulting in respective reductions in memory consumption of 14\% and 15\%.
}

\section{Related Work}

\noindent{\bf Operator fusion and layout optimizations.}
Operator fusion is a commonly used optimization technique in many advanced DNN inference frameworks. 
In the past, MNN~\cite{Ali-MNN}, TFLite~\cite{TensorFlow-Lite}, NCNN~\cite{Ni_ncnn_2017}, and Pytorch-Mobile~\cite{paszke2019pytorch} have all  employed fixed-pattern fusion, which is based on identifying specific  operator combinations to be fused (and 
thus limiting fusion for those cases). 
TVM~\cite{chen2018tvm} recently started supporting operator fusion with more general rules by classifying the operator into three categories~\cite{tvmconvertlayout}: 1) Layout agnostic (e.g., {\tt ReLU}), 2) Lightly-layout sensitive (e.g., {\tt Reduce}), and  3) Heavily-layout sensitive (e.g., {\tt Conv}). 
DNNFusion~\cite{niu2021dnnfusion}  generates fusion plans by classifying operators and their combinations, allowing for a wider range of fusion optimizations.
However, it cannot eliminate explicit data transformation operators through improved layouts.
Furthermore, with our Layout Transformation Elimination, \projectname offers even more opportunities for operator fusion compared to DNNFusion.
Our evaluation demonstrates superior results in terms of fusion rate and latency.
Ivanov et al.~\cite{ivanov2021data} aim to reduce data movement in Transformer training on desktop GPUs, and 
similar to TVM, they propose an optimization framework that systematically fuses operators and selects data layouts by classifying the operator into three categories. 
Different types of operators can only be fused if they have the exact same iteration space, except for the reduction dimension, including both the order and size of the dimensions. This approach cannot optimize (including 
eliminating layout transformations) 
cases involving frequent changes in dimension order and size, which  {\tt Reshape} and {\tt Transpose} kind
of operations explicitly do. 

\noindent{\bf DNN inference frameworks on mobile.}
As AI applications on mobile devices continue to grow, there is a strong emphasis on optimizing DNN inference frameworks for mobile platforms.
Efforts such as DeepEar~\cite{lane2015deepear}, DeepSense~\cite{yao2017deepsense}, Mobisr~\cite{lee2019mobisr}, and others ~\cite{tan2021efficient, han2016mcdnn,lane2016deepx,huynh2017deepmon,xu2018deepcache},
have primarily focused on specific tasks or traditional convolutional neural networks, or require special hardware support.
There are other research directions that target DNN scheduling for heterogeneous platforms, 
including Band ~\cite{jeong2022band}, BlastNet~\cite{ling2022blastnet}, CoDL~\cite{jia2022codl}, and others ~\cite{Xiang2019pipeline,sung2023decentralized}. 

\noindent{\bf Other DNN execution optimization frameworks.}
The advancement of Artificial General Intelligence (AGI) has been significantly bolstered by the evolution of Transformer-based models, paralleled by notable progress in supporting acceleration frameworks like 
FlashAttention~\cite{dao2022flashattention, dao2023flashattention}, and vLLM~\cite{kwon2023pageattention}. 
FlashAttention improves the attention mechanism in Transformers by using tiling strategies to minimize memory access between the GPU's high bandwidth memory (HBM) and on-chip SRAM. 
However, FlashAttention requires extensive register usage in its kernel, which limits its effectiveness on mobile GPUs with limited register capacity ~\cite{chen2023speed}. 
\projectname focuses on Layout Transformation Elimination,  essentially complementing  their work  -- 
combining two approaches could be an interesting future direction. 
vLLM is an efficient LLMs serving system targeting memory fragmentation issues,  inspired by the classical virtual memory and paging techniques used in operating systems.
vLLMs proposes PagedAttention to enhance processing throughput and capability for long sequences in LLMs, achieving near-zero waste in KV cache memory.
TensorFlow-XLA~\cite{tensorflow-xla} and TorchInductor~\cite{TorchInductor2022} are advanced compilers that translate computational graphs and linear algebra expressions into machine code. TensorFlow-XLA uses a ``pad-reshape-transpose'' strategy to optimize layout transformation and tiling, which can lead to additional overhead for layout conversion. 
On the other hand, TorchInductor relies on pre-assigned layouts of specific operators or satisfies layout constraints from library calls (e.g., TensorRT). 
\section{Discussion and Future Work}

\noindent\textbf{\revisioncr{Automating operator categorization.}}
The classification of operators within \projectname is designed to evaluate their sensitivity to variations in both computational pattern 
and data access pattern (i.e., computation performance and output layout). 
By utilizing established intermediate representations like tensor expressions supported by TVM, 
it is possible to create a tool that automates and adapts for operator categorization.
This approach could be further enhanced through the integration of symbolic manipulation tools, enabling the handling of more intricate scenarios. 
Our future work will further enhance this process and explore the potential of automating the categorization of operators.

\noindent\textbf{\revisioncr{Layout optimizations beyond mobile devices.}}
\projectname optimizes data layout transformations commonly seen in modern DNN computational graphs, 
leading to significant performance improvements across various DNNs, including LLMs.
The optimizations in \projectname, such as Layout Transformation Elimination and layout selection,
are also applicable across different platforms, including desktop-level GPUs.
However, mobile GPUs benefit more from \projectname.
This is because mobile GPUs have more restricted memory size and bandwidth, 
which makes their performance more sensitive to data transformation elimination optimizations.
In addition, mobile GPUs rely on an efficient 2.5D texture memory for data processing, 
which offers us further data layout optimization opportunities. 
Implementations on desktop-level GPUs mainly rely on shared memory and cache, and their texture memory usage is relatively restricted.
\projectname also has the potential to be extended to other platforms, such as FPGAs and ASICs, 
where the memory hierarchy and data layout optimizations are also critical for performance.

\section{Conclusions}
This paper introduces \projectname, a novel framework to significantly improve the performance of DNN execution on mobile devices for both CNN and Transformer structures. 
\projectname categorizes DNN operators into four groups based on their input/output layouts and computations, considers combinations of producer-consumer edges between the operators, and searches optimized layouts with multiple carefully designed methods. 
Our extensive experiments with 18 cutting-edge models show significant speedup compared to MNN, NCNN, TFLite, TVM, and DNNFusion.
Looking ahead, this work paves the way for further research in optimizing LLMs (with larger parameters and more operators) on mobile devices. 
Considering the nature of mobile GPUs (e.g., limited memory bandwidth and register file size), future work of \projectname will 
explore combining different optimizations (e.g., pruning, quantization)
that dynamically adapt to the ever-changing landscape of AGI and mobile hardware. 
\begin{acks}
\revisioncr{
The authors want to extend their appreciation to the anonymous reviewers and shepherd, Tim Harris, for their valuable and thorough feedback. 
All of these constructive suggestions have greatly contributed to enhancing this paper. 
This work was supported in part by the National Science Foundation (NSF) under the awards of 
CCF-2047516 (CAREER), CCF-2146873, CCF-2333895, CCF-2334273,  
CNS-2312207, CNS-2230944,  CNS-2341378,
III-2008557, 
OAC-2333899,
and DOE DE-SC0021293.
Any errors and opinions are not those of the NSF and are attributable solely to the author(s). 
The authors also acknowledge William \& Mary Research Computing for providing computational resources.
}
\end{acks}

\balance
\bibliographystyle{plain}
\bibliography{reference}

\end{document}